\documentclass[11pt]{article}
\usepackage[hyperref]{acl2020}
\usepackage{times}
\usepackage{latexsym}
\usepackage{soul}
\usepackage{amsmath}
\usepackage{bm}
\usepackage{enumitem}
\usepackage{amssymb}
\usepackage{booktabs}
\usepackage{graphicx}
\usepackage{multirow}
\usepackage{fixltx2e}
\usepackage{url}

\aclfinalcopy

\newcommand{\link}{l}

\newcommand{\coarse}{c}
\newcommand{\dummy}{$\epsilon$}

\newcommand{\best}[1]{\textbf{#1}}
\newcommand{\F}{$\rm F_1$}
\newcommand{\datasetname}{RAMS}

\newcommand{\ont}[1]{\texttt{#1}}
\newcommand{\role}[3][0]{\underset{\textsc{#3}}{\underline{\textrm{#2}}}^{#1}}
\newcommand{\doc}{$\mathcal{D}$}
\newcommand{\ramsbest}{68.3}
\newcommand{\ramsbestconstrained}{73.3}

\title{Multi-Sentence Argument Linking}

\author{Seth Ebner\thanks{~~Equal Contribution}~~~ Patrick Xia\footnotemark[1]~~~ Ryan Culkin~~~ Kyle Rawlins~~~ Benjamin Van Durme \\
  Johns Hopkins University\\
  {\tt \{seth, paxia\}@cs.jhu.edu} \\
  {\tt \{rculkin, kgr, vandurme\}@jhu.edu}
}

\date{}

\begin{document}
\maketitle
\begin{abstract}
  We present a novel document-level model for finding argument spans that fill an event's roles, connecting related ideas in sentence-level semantic role labeling and coreference resolution. Because existing datasets for cross-sentence linking are small, development of our neural model is supported through the creation of a new resource, {\bf R}oles {\bf A}cross {\bf M}ultiple {\bf S}entences (\datasetname{}), which contains 9,124 annotated events across 139 types. We demonstrate strong performance of our model on \datasetname{} and other event-related datasets.\footnote{Data and code at \url{http://nlp.jhu.edu/rams/}.}
\end{abstract}

\section{Introduction}

Textual event descriptions may span multiple sentences, yet large-scale datasets predominately annotate for events and their arguments at the sentence level. This has driven researchers to focus on sentence-level tasks such as semantic role labeling (SRL), even though perfect performance at such tasks would still enable a less than complete understanding of an event at the document level.

In this work, we approach event understanding as a form of \emph{linking}, more akin to coreference resolution than sentence-level SRL. An event trigger \emph{evokes} a set of roles regarded as latent arguments, with these implicit arguments then potentially linked to explicit mentions in the text.

\begin{figure}[t]
\includegraphics[scale=0.195]{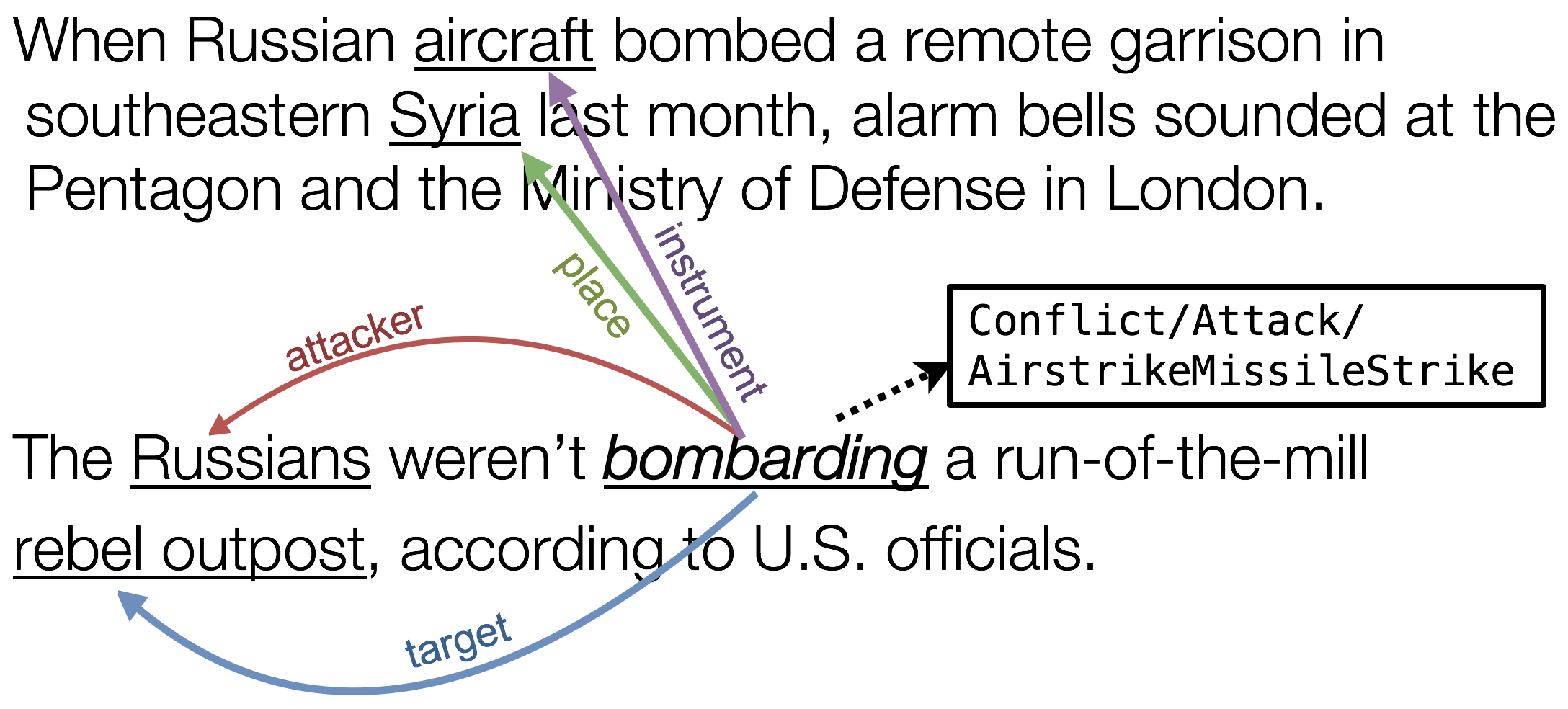}
\caption{A passage annotated for an event's \ont{type}, \textbf{\textit{\underline{trigger}}}, and \underline{arguments}. Each arc points from the trigger to the argument that fills the labeled role.}
\label{fig: exm}
\end{figure}

Consider the example in \autoref{fig: exm}: the \ont{Air\-strike\-Missile\-Strike} event (triggered by ``bombarding'') gives rise to a frame or set of type-level roles (\ont{attacker}, \ont{target}, \ont{instrument}, \ont{place}) with the referents (``Russians'', ``rebel outpost'', ``aircraft'', ``Syria'').\footnote{\dummy~would indicate there is no explicit referent in the text.} Intuitively we recognize the possible existence of fillers for these roles, for example, the \ont{place} of the particular \ont{Air\-strike\-Missile\-Strike} event. These implicit arguments are linked to explicit arguments in the document (i.e., text spans). We refer to the task of finding explicit argument(s) to fill each role for an event as \emph{argument linking}.

Prior annotation of cross-sentence argument links has produced small datasets, with a focus either on a small number of predicate types~\cite{gerber-chai-2010-beyond, gerber-chai-2012-semantic, feizabadi-pado-2014-crowdsourcing} or on a small number of documents~\cite{ruppenhofer-etal-2010-semeval}. To enable the development of a neural model for argument linking, we produce \textbf{R}oles \textbf{A}cross \textbf{M}ultiple \textbf{S}entences (\datasetname{}), a dataset of 9,124 annotated events from news based on an ontology of 139 event types and 65 roles. In a 5-sentence window around each event trigger, we annotate the closest argument span for each role.

Our model builds on recent ideas in span selection models \cite{lee-etal-2018-higher, he-etal-2018-jointly, ouchi-etal-2018-span}, used in this work for the multi-sentence argument linking task for \datasetname{} and for several other event-based datasets \cite[AIDA Phase 1]{gerber-chai-2012-semantic, pradhan-etal-2013-towards, pavlick-etal-2016-gun}. On \datasetname{} our best model achieves \ramsbest{}~\F{}, and it achieves \ramsbestconstrained{}~\F{} when event types are also known, outperforming strong baselines. We also demonstrate effective use of \datasetname{} as pre-training for a related dataset.

Our main contributions are a novel model for argument linking and a new large-scale dataset for the task. Our dataset is annotated for arguments across multiple sentences and has broader coverage of event types and more examples than similar work. Our experiments highlight our model's adaptability to multiple datasets. Together, these contributions further the automatic understanding of events at the document level.

\section{Non-local Arguments}

We are not the first to consider non-local event arguments; here we review prior work and refer to \citet{OGorman19} for further reading. Whereas local (sentence-level) event arguments are well-studied as semantic role labeling---utilizing large datasets such as OntoNotes~5.0 \cite{weischedel2013ontonotes, pradhan-etal-2013-towards}---existing datasets annotated for non-local arguments are too small for training neural models.

Much of the effort on non-local arguments, sometimes called \emph{implicit} SRL, has focused on two datasets: SemEval-2010 Task 10~\cite{ruppenhofer-etal-2010-semeval} and Beyond NomBank (henceforth \textbf{BNB}) \cite{gerber-chai-2010-beyond, gerber-chai-2012-semantic}. These datasets are substantially smaller than \datasetname{}: the SemEval Task 10 training set contains 1,370 frame instantiations over 438 sentences, while BNB contains 1,247 examples covering just 10 nominal predicate types. Multi-sentence AMR (MS-AMR) \cite{ogorman-etal-2018-amr, knight2020amr} contains 293 documents annotated with a document-level adaptation of the Abstract Meaning Representation (AMR) formalism. \citet{OGorman19} notes that the relatively small size of the MS-AMR and SemEval datasets hinders supervised training. In contrast to these datasets, \datasetname{} contains 9,124 annotated examples covering a wide range of nominal and verbal triggers.

\begin{figure}[t]
\centering{\includegraphics[scale=0.55]{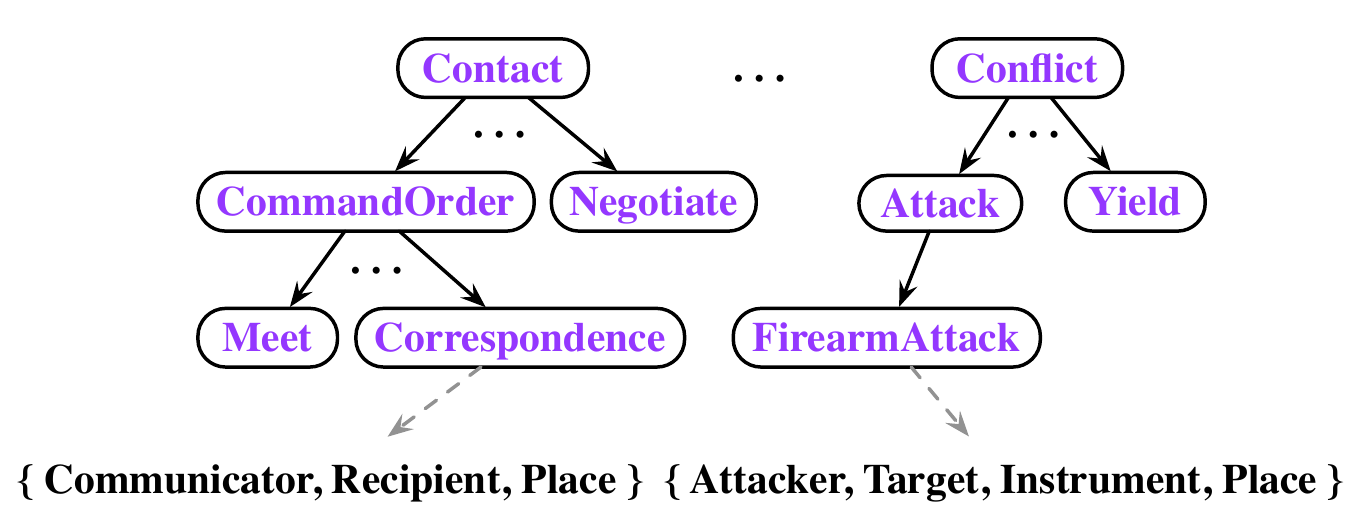}}
\caption{Subset of the AIDA-1 ontology illustrating the three-level \ont{Type/Subtype/Sub-subtype} event hierarchy. Dashed gray edges point to roles for two event nodes, which have one role in common (\ont{Place}).}
\label{fig:hierarchy}
\end{figure}

Under the DARPA AIDA program, the Linguistic Data Consortium (LDC) has annotated document-level event arguments under a three-level hierarchical event ontology (see \autoref{fig:hierarchy}) influenced by prior LDC-supported ontologies such as ERE and ACE. These have been packaged as the AIDA Phase 1 Practice\footnote{LDC2019E04 (data); LDC2019E07 (annotations)} and Eval\footnote{LDC2019E42 (data); LDC2019E77 (annotations)} releases (henceforth \textbf{AIDA-1}), currently made available to performers in the AIDA program and participants in related NIST evaluations.\footnote{While rarely freely released, historically such collections are eventually made available under a license to anyone, under some timeline established within a program.}  AIDA-1 documents focus on recent geopolitical events relating to interactions between Russia and Ukraine. Unless otherwise noted, statistics about AIDA-1 pertain only to the Practice portion of the dataset.

For each document in LDC's collection, only AIDA-salient events are annotated. This protocol does not guarantee coverage over the event ontology: 1,559 event triggers are annotated in the text portion of the collection, accounting for only 88 of the 139 distinct event sub-subtypes in the ontology. Our dataset, \datasetname{}, employs the same annotation ontology but is substantially larger and covers all 139 types in the ontology. \autoref{fig:type-coverage-frequency} (\S\ref{sec:rams}) compares the two datasets.

Across multiple datasets, a substantial number of event arguments are observed to be non-local. For example, \citet{gerber-chai-2012-semantic} found that their annotation of non-local arguments added 71\% (relative) role coverage to NomBank annotations. Additionally, 38.1\% of the annotated events in AIDA-1 have an argument outside the sentence containing the trigger. This phenomenon is not surprising in light of the analysis of zero anaphora and definite null complements by~\citet{fillmore1986pragmatically} and the distinction between ``core'' and ``non-core'' frame elements or roles in FrameNet~\cite{baker-etal-1998-berkeley-framenet} and PropBank~\cite{palmer-etal-2005-proposition}.

As previous datasets have been small, various approaches have been taken to handle scarcity. To obtain more training data, \citet{silberer-frank-2012-casting} created artificial instances from data annotated jointly for coreference and semantic roles. \citet{roth-frank-2013-automatically} automatically induced implicit arguments from pairs of comparable texts, but recovered a proportionally small set of additional arguments. \citet{feizabadi-pado-2015-combining} combined existing corpora to increase and diversify sources of model supervision. \citet{cheng-erk-2018-implicit, cheng2019implicit} approached the data scarcity problem by recasting implicit SRL as a cloze task and as a reading comprehension task, for which data can be generated automatically. 

The TAC KBP event argument extraction task also seeks arguments from document contexts. However, in our work we are concerned with reified events (explicit mentions) and links between event mentions and argument mentions rather than entity-level arguments (coreference clusters).

\section{\datasetname{}}\label{sec:rams}

Motivated by the scarcity of data for training neural models to predict non-local arguments, we constructed \textbf{R}oles \textbf{A}cross \textbf{M}ultiple \textbf{S}entences (\datasetname{}), a crowd-sourced dataset with annotations for 9,124 events following the AIDA ontology. We employed the AIDA ontology in \datasetname{} so-as to be most similar to an existing corpus already being investigated by various members of the community. Each example consists of a typed \textit{trigger} span and 0 or more \textit{argument} spans in an English document. A trigger span is a word or phrase that evokes a certain event type in context, while argument spans denote role-typed participants in the event (e.g., the \ont{Recipient}). Trigger and argument spans are token-level $[start, end]$ offsets into a tokenized document. 

Typically, event and relation datasets annotate only the argument spans that are in the same sentence as the trigger, but we present annotators with a \textit{multi-sentence} context window surrounding the trigger. Annotators may select argument spans in any sentence in the context window. 

\subsection{Dataset Description}

\paragraph{Data Source} We used Reddit, a popular internet forum, to filter a collection of news articles to be topically similar to AIDA-1. After applying a set of criteria based on keywords, time period, and popularity (listed in Appendix \ref{appendix:data:reddit}) we identified approximately 12,000 news articles with an average length of approximately 40 sentences.

\paragraph{Annotation} We manually constructed a mapping from each event ((sub-)sub)type to a list of lexical units (LUs) likely to evoke that type.\footnote{For example, \ont{Conflict/Attack/SetFire} is evoked by \emph{inferno}, \emph{blaze}, and \emph{arson} (and word forms).} This mapping was designed to give high precision and low recall, in that for a given (\texttt{Type}, \texttt{LUs}) pair, the items in \texttt{LUs} are all likely to evoke the \texttt{Type}, although \texttt{LUs} can omit items that also evoke the \texttt{Type}. On average, $|\texttt{LUs}| = 3.9$.

We performed a soft match\footnote{We stem all words and ignore case.} between every LU and every word in our text collection to select candidate sentences for each event type. This matching procedure produced approximately 94,000 candidates, which we balanced by sampling the same number of sentences for each LU.

Candidate sentences were then vetted by crowd-sourcing to ensure that they evoked their associated event type and had positive factuality. We collected judgments on approximately 17,500 candidate sentences, of which 52\% were determined to satisfy these constraints, yielding 9,124 sentences containing a LU trigger. Using these sentences we then collected multi-sentence annotations, presenting annotators with a 5-sentence window containing two sentences of context before the sentence with the trigger and two sentences after.\footnote{If fewer than two sentences appeared before/after the trigger, annotators were shown as many sentences as were available.} Annotators then selected in the context window a span to fill each of the event's roles.

\begin{table}[t]
\small
\centering
\begin{tabular}{lcccc}
\toprule
 & Train & Dev & Test & Total\\
\midrule
Docs & 3,194 & 399 & 400 & 3,993 \\
Examples & 7,329 & 924 & 871 & 9,124 \\
Event Types & 139 & 131 & -- & 139 \\
Roles & 65 & 62 & -- & 65 \\
Arguments & 17,026 & 2,188 & 2,023 & 21,237 \\
\bottomrule
\end{tabular}
\caption{Sizes and coverage of \datasetname{} splits. \datasetname{} covers all of the 139 event types and 65 roles types in the AIDA Phase~1 ontology.}
\label{tab:ramsstatistics}
\end{table}

A window size of five sentences was chosen based on internal pilots and supported by our finding that 90\% of event arguments in AIDA-1 are recoverable in this window size. Similarly, \citet{gerber-chai-2010-beyond} found that in their data almost 90\% of implicit arguments can be resolved in the two sentences preceding the trigger.\footnote{Arguments following the trigger were not annotated.} Arguments fall close to the trigger in \datasetname{} as well: 82\% of arguments occur in the same sentence as the trigger. On average, we collected 66 full annotations (trigger and arguments) per event type. \autoref{tab:ramsstatistics} shows dataset size and coverage. All aspects of the protocol, including the annotation interface and instructions, are included in Appendix \ref{appendix:ramsdata}.

\paragraph{Inter-Annotator Agreement}
We randomly selected 93 tasks for redundant annotation in order to measure inter-annotator agreement, collecting five responses per task from distinct users. 68.5\% of the time, all annotators mark the role as either absent or present. Less frequently (21.7\%), four of the five annotators agree, and rarely (9.8\%) is there strong disagreement.

We compute pairwise agreement for span boundaries. For each annotated (event, role) combination, we compare pairs of spans for which both annotators believe the role is present. 55.3\% of the pairs agree exactly. Allowing for a fuzzier match, such as to account for whether one includes a determiner, spans whose boundaries differ by one token have a much higher agreement of 69.9\%. Fewer spans agree on the start boundary (59.8\%) than on the end (73.5\%), while 78.0\% match at least one of the two boundaries. We demonstrate data quality in \S\ref{sec:rams:aida} by showing its positive impact on a downstream task.

\begin{figure}[h]
\centering
\includegraphics[width=0.9\linewidth]{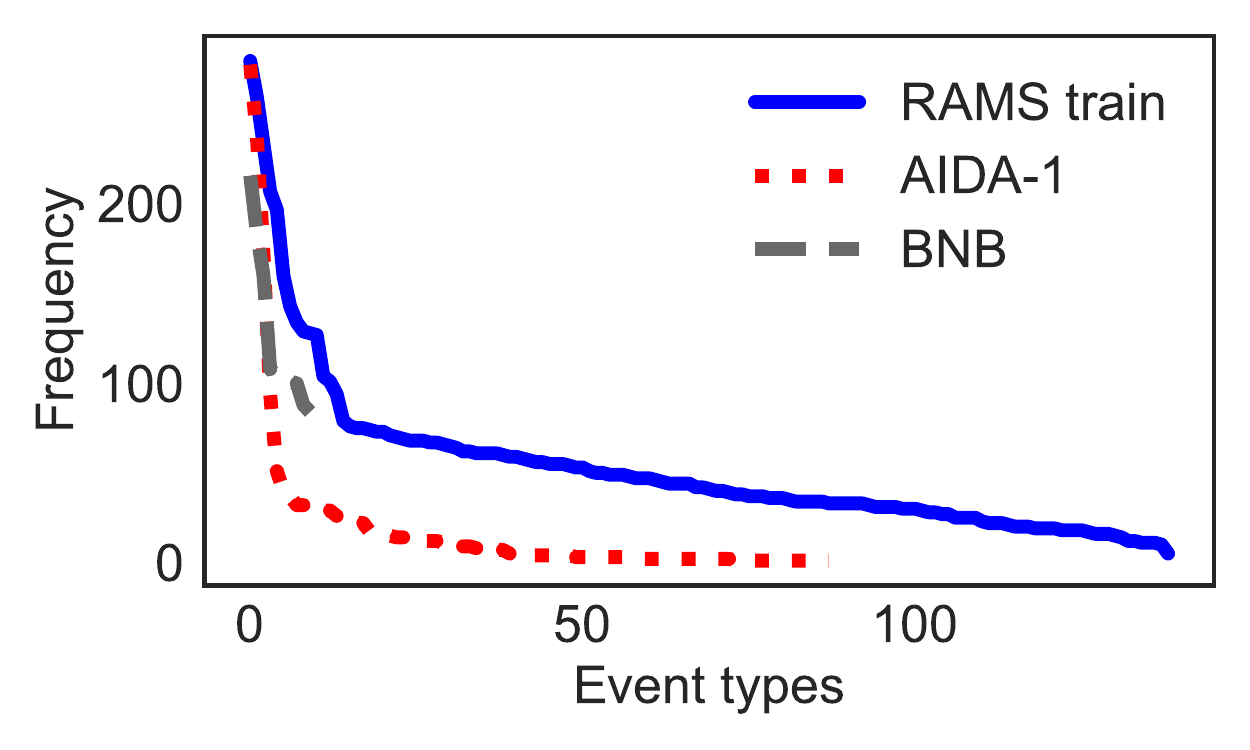}
\caption{Comparison of frequency of event types in various datasets sorted by decreasing frequency in that dataset. \datasetname{} has a heavier tail than AIDA-1 and BNB and broader coverage of events.}
\label{fig:type-coverage-frequency}
\end{figure}

\paragraph{Comparisons to Related Datasets}
Comparisons of event type coverage among \datasetname{}, AIDA-1, and BNB ~\cite{gerber-chai-2010-beyond,gerber-chai-2012-semantic} are given in \autoref{fig:type-coverage-frequency}. \datasetname{} provides larger and broader coverage of event types than do AIDA-1 and BNB. By design, BNB focuses on only a few predicate types, but we include its statistics for reference. More figures regarding type and role coverage are included in Appendix \ref{appendix:data:coverage}.

\paragraph{Related Protocols} \citet{feizabadi-pado-2014-crowdsourcing} also considered the case of crowdsourcing annotations for cross-sentence arguments. Like us, they provided annotators with a context window rather than the whole document, annotating two frames each with four roles over 384 predicates. Annotators in that work were shown the sentence containing the predicate and the three previous sentences, unlike ours which shows two preceding and two following sentences.

Rather than instructing annotators to highlight spans in the text (``marking''), \citet{feizabadi-pado-2014-crowdsourcing} directed annotators to fill in blanks in templatic sentences (``gap filling''). We in contrast require annotators to highlight mention spans directly in the text. 

Our protocol of event type verification followed by argument finding is similar to the protocol supported by interfaces such as SALTO~\cite{burchardt-etal-2006-salto} and that of~\citet{fillmore-etal-2002-framenet}.

\section{Model}

We formulate \textit{argument linking} as follows, similar to the formulation in \citet{das-etal-2010-probabilistic}. Assume a document \doc{} contains a set of described events $\mathcal{E}$, each designated by a trigger---a text span in \doc{}. The type of an event $e$ determines the set of roles the event's arguments may take, denoted $\mathcal{R}_e$. For each $e \in \mathcal{E}$, the task is to link the event's roles with arguments---text spans in \doc{}---if they are attested. Specifically, one must find for each $e$ all $(r, a)$ pairs such that $r \in \mathcal{R}_e$ and $a \in \mathcal{D}$. This formulation does not restrict each role to be filled by only one argument, nor does it restrict each explicit argument to take at most one role.

\subsection{Architecture}\label{sec:arch}

Our model architecture is related to recent models for SRL~\cite{he-etal-2018-jointly, ouchi-etal-2018-span}. Contextualized text embeddings are used to form candidate argument span representations, $\mathcal{A}$. These are then pruned and scored alongside the trigger span and learned role embeddings to determine the best argument span (possibly none) for each event and role, i.e., $\text{argmax}_{a \in \mathcal{A}} P(a \mid e, r)$ for each event $e \in \mathcal{E}$ and role $r \in \mathcal{R}_e$.

\paragraph{Representations} 
To represent text spans, we adopt the convention from~\citet{lee-etal-2017-end} that has been used for a broad suite of core NLP tasks \cite{swayamdipta-etal-2018-syntactic, he-etal-2018-jointly, tenney2018what}. A bidirectional LSTM encodes each sentence's contextualized embeddings~\cite{peters-etal-2018-deep, Devlin18BERT}. The hidden states at the start and end of the span are concatenated along with a feature vector for the size of the span and a soft head word vector produced by a learned attention mask over the word vectors (GloVe embeddings \cite{pennington-etal-2014-glove} and character-level convolutions) within the span.

We use this method to form representations of trigger spans, $\bf{e}$, and of candidate argument spans, $\bf{a}$. We learn a separate embedding, $\bf{r}$, for each role in the ontology, $r \in \mathcal{R}$. Since our objective is to link candidate arguments to event-role pairs, we construct an event-role representation\footnote{As a role for an event evokes an \textit{implicit discourse referent}, this can be regarded as an implicit discourse referent representation.} by applying a feed-forward neural network ($\text{F}_{\tilde{a}}$) to the event trigger span and role embedding:
\begin{align}
    \tilde{\bf{a}}_{e,r} = \text{F}_{\tilde{a}}([\bf{e};\bf{r}]) \label{eqn:trigrole}
\end{align}
 
This method is similar to one for forming edge representations for cross-sentence relation extraction \cite{song-etal-2018-n}, but contrasts with prior work which limits the interaction between $r$ and $e$ \cite{he-etal-2018-jointly, tenney2018what}.

\paragraph{Pruning}
Given a document with $n$ tokens, there are $O(n^2)$ candidate argument text spans, which leads to intractability for large documents.  Following \citet{lee-etal-2017-end} and \citet{he-etal-2018-jointly}, we consider within-sentence spans up to a certain width (giving $O(n)$ spans) and score each span, $a$, using a learned unary function of its representation: $s_A(a) = \textbf{w}_A^\top\text{F}_A (\textbf{a})$. We keep the top $\lambda_A n$ spans ($\lambda_A$ is a hyperparameter) and refer to this set of high-scoring candidate argument spans as $\mathcal{A}$.

In an unpruned model, we need to create at least $\sum_{e}|\mathcal{R}_{e}|$ event-role representations and evaluate $\Omega(n\sum_{e}|\mathcal{R}_{e}|)$ combinations of events, roles, and arguments, which can become prohibitively large when there are numerous events and roles. Assuming the number of events is linear in document length, the number of combinations would be quadratic in document length (rather than quadratic in sentence length as in \citet{he-etal-2018-jointly}).

\citet{lee-etal-2018-higher} addressed this issue in coreference resolution, a different document-level task, by implementing a coarse pruner to limit the number of candidate spans that are subsequently scored. For our model, any role can potentially be filled (if the event type is not known). Thus, we do not wish to prematurely prune $(e, r)$ pairs, so we must further prune $\mathcal{A}$. Rather than scoring $a \in \mathcal{A}$ with every event-role pair $(e, r)$, we assign a score between $a$ and every event $e$. This relaxation reflects a loose notion of how likely an argument span is to participate in an event, which can be determined irrespective of a role: 
\begin{align}
    s_\coarse(e, a) = {\bf e}^\top \textbf{W}_\coarse{\bf a} + s_A(a) + s_{E}(e) + \phi_\coarse(e, a) \nonumber
\end{align}
\noindent where $\textbf{W}_\coarse$ is learned and $\phi_\coarse(e,a)$ are task-specific features. We use $\mathcal{A}_e \subseteq \mathcal{A}$ to refer to the top-$k$-scoring candidate argument spans in relation to $e$. 

\paragraph{Scoring}
We introduce a link scoring function, $\link(a, \tilde{a}_{e,r})$, between candidate spans $a \in \mathcal{A}_e$ and event-role pairs $\tilde{a}_{e,r} = (e, r) \in \mathcal{E} \times \mathcal{R}$.\footnote{If the type of $e$ is known, then we could restrict $r \in \mathcal{R}_e$.} The scoring function decomposes as:
\begin{align}
    \link(a, \tilde{a}_{e,r}) &= s_{E,R}(e, r) + s_{A,R}(a, r) \nonumber \\&+ s_\link(a, \tilde{a}_{e,r}) + s_\coarse(e, a),~~~a \not= \epsilon \label{eqn:link}\\
    s_{E}(e) &= \textbf{w}_{E}^\top\text{F}_{E} (\textbf{e}) \nonumber \\
    s_{E,R}(e, r) &= \textbf{w}_{E,R}^\top\text{F}_{E,R} ([\textbf{e};\textbf{r}]) \nonumber \\
    s_{A,R}(a, r) &= \textbf{w}_{A,R}^\top\text{F}_{A,R} ([\textbf{a};\textbf{r}]) \nonumber\\
    s_\link(a, \tilde{a}_{e,r}) &= \textbf{w}_\link^\top\text{F}_\link([\textbf{a}; \tilde{\textbf{a}}_{e,r};  \textbf{a} \circ \tilde{\textbf{a}}_{e,r}; \nonumber\\ & ~~~~~~~~~~~~~~~~~\phi_\link(a, \tilde{a}_{e,r})]) \label{eqn:big}
\end{align}
where $\phi_\link(a,\tilde{a}_{e,r})$ is a feature vector containing information such as the (bucketed) token distance between $e$ and $a$.\footnote{Distance $= {\max(e_{start} - a_{end}, a_{start} - e_{end})}$.} $\text{F}_{x}$ are feed-forward neural networks, and $\textbf{w}_{x}$ are learned weights. The decomposition is inspired by \citet{lee-etal-2017-end} and \citet{he-etal-2018-jointly}, while the direct scoring of candidate arguments against event-role pairs, $s_\link(a, \tilde{a}_{e,r})$, bears similarities to the approach taken by~\citet{schenk-chiarcos-2016-unsupervised}, which finds the candidate argument whose representation is most similar to the prototypical filler of a frame element (role). 

\paragraph{Learning} We denote ``no explicit argument'' by $\epsilon$ and assign it link score $\link(\epsilon, \tilde{a}_{e, r}) \triangleq 0$, which acts as a threshold for the link function. For every event-role-argument triple $(e, r, a)$, we maximize
\begin{align*}
P(a \mid e, r) = \frac{\exp\{\link(a, \tilde{a}_{e,r})\}}{\sum_{a' \in \mathcal{A}_e \cup \{\epsilon\}}\exp{\{\link(a', \tilde{a}_{e,r})\}}}.
\end{align*}

\paragraph{Decoding} We experiment with three decoding strategies: \textit{argmax}, \textit{greedy}, and \textit{type-constrained}. If we assume each role is satisfied by exactly one argument (potentially $\epsilon$), we can perform \textit{argmax} decoding independently for each role:
\begin{align*}
\hat{a} = \text{argmax}_{a \in \mathcal{A}_e \cup \{\epsilon\}} P(a \mid e,r)
\end{align*}

To instead predict multiple non-overlapping arguments per role, we could use ${P(\epsilon \mid e,r)}$ as a threshold in \textit{greedy} decoding \cite{ouchi-etal-2018-span}.

We may know the gold event types and the mapping between events $e$ and their permitted roles, $\mathcal{R}_e$. While this information can be used during training, we take a simpler approach of using it for \textit{type-constrained} decoding (TCD). If an event type allows $m_{r}$ arguments for role $r$, we keep only the top-scoring $m_r$ arguments based on link scores.

\subsection{Related Models}

Our model is inspired by several recent span selection models \cite{he-etal-2018-jointly, lee-etal-2018-higher, ouchi-etal-2018-span}, as well as the long line of neural event extraction models~\citep[\emph{inter alia}]{chen-etal-2015-event, nguyen-etal-2016-joint-event}. \citet{OGorman19} speculates a joint coreference and SRL model in which implicit discourse referents are generated for each event predicate and subsequently clustered with the discovered referent spans using a model for coreference, which is similar to the approach of \citet{silberer-frank-2012-casting}. \citet{OGorman19} further claims that span selection models would be difficult to scale to the document level, which is the regime we are most interested in. We focus on the implicit discourse referents (i.e., the event-role representations) for an event and link them to argument mentions, rather than cluster them using a coreference resolution system or aggregate event structures across multiple events and documents \cite{wolfe-etal-2015-predicate}. Our approach is also similar to the one used by \citet{das-etal-2010-probabilistic} for FrameNet parsing.

\paragraph{CoNLL 2012 SRL}
As our model bears similarities to the SRL models proposed by \citet{he-etal-2018-jointly} and \citet{ouchi-etal-2018-span}, we evaluate our model on the sentence-level CoNLL 2012 dataset as a sanity check. Based on a small hyperparameter sweep, our model achieves 81.4~\F{} when given gold predicate spans and 81.2~\F{} when not given gold predicates.\footnote{We use ELMo \cite{peters-etal-2018-deep} in these experiments. \citet{he-etal-2018-jointly} achieve 85.5~\F{} with gold predicates and 82.9~\F{} without gold predicates, and \citet{ouchi-etal-2018-span} achieve 86.2~\F{} with gold predicates.} Our model's recall is harmed because our span pruning occurs at the document level rather than at the sentence level, which leads to overpruning in some sentences. Although our model is designed to accommodate cross-sentence links, it maintains competitive performance on sentence-level SRL.

\section{RAMS Experiments and Results}\label{sec:ramsexps}

In the following experiments, for each event the model is given the (gold) trigger span and the (gold) spans of the arguments. The model finds for each role the best argument(s) to fill it. Predictions are returned as trigger-role-argument triples.

We use feature-based BERT-base \cite{Devlin18BERT}---mixing layers 9 through 12---by splitting the documents into segments of size 512 subtokens and encoding each segment separately.\footnote{0.2\% of the training documents span multiple segments.}

We perform preliminary sweeps across hyperparameter values, which are then fixed while we perform a more exhaustive sweep across scoring features. We also compare argmax decoding with greedy decoding during training. The best model is selected based on \F{} on the development set, and ablations are reported in \autoref{tab:ramsablations}. Our final model uses greedy decoding, $s_{A,R}$, and $s_l$ and omits $s_{E,R}$ and $s_\coarse$ (see \autoref{eqn:link}). More details can be found in \autoref{appendix:rams}.

\begin{table}
\small
\centering
\begin{tabular}{lcccc}
\toprule
Model & Dev. \F{} & P & R & \F{} \\
\midrule
Our model & 69.9 & 62.8 & \best{74.9} & 68.3 \\ 
Our model\textsuperscript{TCD} & \best{75.1} &  78.1 & 69.2 & \best{73.3} \\
\midrule
Most common & 17.3 & 15.7 & 15.7 & 15.7 \\
Fixed trigger\textsuperscript{TCD} & 60.2 & \best{83.7} & 41.9  & 55.8 \\ 
Context as trigger\textsuperscript{TCD} & 62.1 & 80.5 & 45.8 & 58.4 \\
\midrule
Distractor arguments & 24.3 & 60.5 & 15.1 & 24.2 \\ 
Distractor arguments\textsuperscript{TCD} & 24.2 & 68.8 & 14.3 & 23.7 \\
No given arguments & 8.7 & 20.2 & 3.5 & 6.0 \\ 
No given arguments\textsuperscript{TCD} & 8.4 & 26.6 & 3.1 & 5.5 \\
\bottomrule
\end{tabular}
\caption{P(recision), R(ecall), and \F{} on \datasetname{} development and test data. TCD designates the use of ontology-aware type-constrained decoding.}
\label{tab:ramsresults}
\end{table}

The results using our model with greedy decoding and TCD are reported in \autoref{tab:ramsresults}. We also report performance of the following baselines: 1) choosing for each link the most common role (\ont{place}), 2) using the same fixed trigger representation across examples, and 3) using the full context window as the trigger. Additionally, we experiment with two other data conditions: 1) linking the correct argument(s) from among a set of distractor candidate arguments provided by a constituency parser~\cite{kitaev-klein-2018-constituency},\footnote{We take as the distractor arguments all (potentially overlapping) \texttt{NP}s predicted by the parser. On average, this yields 44 distractors per training document.} and 2) finding the correct argument(s) from among all possible spans up to a fixed length.

For the distractor experiment, we use the same hyperparameters as for the main experiment. When not given gold argument spans, we consider all spans up to 5 tokens long and change only the hyperparameters that would prune less aggressively. We hypothesize that the low performance in this setting is due to the sparsity of annotated spans compared to the set of all enumerated spans. In contrast, datasets such as CoNLL 2012 are more densely annotated, so the training signal is not as affected when the model must determine argument spans in addition to linking them.

Finally, we examine the effect of TCD to see whether the model effectively uses gold event types if they are given. TCD filters out illegal predictions, boosting precision. Recall is still affected by this decoding strategy because the model may be more confident in the wrong argument for a given role, thus filtering out the less confident, correct one. Nevertheless, using gold types at test time generally leads to gains in performance.

\begin{table}[t]
\small
\centering
\begin{tabular}{lcc}
\toprule
Model & Greedy & TCD \\
\midrule
Our model & \best{69.9} & 75.1 \\ 
- distance score & 69.0 & 74.3 \\ 
- $s_\link(a, \tilde{a}_{e,r})$ & 54.9 & 58.4 \\ 
- $s_{A,R}(a, r)$ & 68.6 & 73.8 \\
+ $s_{E,R}(e, r)$ & 69.5 & 74.4 \\ 
+ $s_\coarse(e, a)$ & 65.9 & 70.6 \\  
\midrule
w/ argmax decoding & \best{69.9} & 75.1 \\ 
\midrule

BERT 6--9 & 69.6 & \best{75.3} \\ 
ELMo & 68.5 & 75.2 \\ 
\bottomrule
\end{tabular}
\caption{\F{} on \datasetname{} dev data when link score components are separately included/excluded~(\autoref{eqn:link}) or other contextualized encoders are used in the best performing model. TCD = type-constrained decoding.}
\label{tab:ramsablations}
\end{table} 

\subsection{Analysis}
\label{sec:rams:analysis}

\paragraph{Ablations}
Ablation studies on development data for components of the link score as well as the contextualized encoder and decoding strategy are shown in~\autoref{tab:ramsablations}. Type-constrained decoding based on knowledge of gold event types improves \F{} in all cases because it removes predictions that are invalid with respect to the ontology.

The most important link score component is the score between a combined event-role and a candidate argument. This result follows intuitions that $s_\link$ is the primary component of the link score since it directly captures the compatibility of the explicit argument and the implicit argument represented by the event-role pair. 

We also experiment with both ELMo \cite{peters-etal-2018-deep} and BERT layers 6--9, which were found to have the highest mixture weights for SRL by \citet{tenney-etal-2019-bert}. We found that BERT generally improves over ELMo and layers 9--12 often perform better than layers 6--9.

\begin{table}[t]
\small
\centering
\begin{tabular}{lccccc}
\toprule
Dist. & \# Gold & \# Predict & P & R & \F{} \\
\midrule
 -2 & 79 (26) & 69 (21) & 81.2 & 70.9 & 75.7 \\ 
 -1 & 164 (33) & 151 (27) & 76.8 & 70.7 & 73.7 \\ 
 ~0 & 1,811 (61) & 1,688 (51) & 77.7 & 72.4 & 75.0 \\ 
 ~1 & 87 (24) & 83 (22) & 78.3 & 74.7 & 76.5 \\ 
 ~2 & 47 (18) & 39 (14) & 87.2 & 72.3 & 79.1 \\
 \midrule
Total & 2,189 (62) & 2,030 (52) & 78.0 & 72.3 & 75.1 \\
\bottomrule
\end{tabular}
\caption{Performance breakdown by distance (number of sentences) between argument and event trigger for our model using TCD over the development data. Negative distances indicate that the argument occurs before the trigger. \# Gold and \# Predict list the number of arguments (and unique roles) at that distance.}
\label{tab:ramsdistancesresults}
\end{table}

\paragraph{Argument--Trigger Distance}
One of the differentiating components of \datasetname{} compared to SRL datasets is its non-local annotation of arguments. At the same time, \datasetname{} uses naturally occurring text so arguments are still heavily distributed within the same sentence as the trigger (\autoref{fig:ramsdistancesstatistics}). This setting allows us to ask whether our model accurately finds arguments outside of the sentence containing the trigger despite the non-uniform distribution. In \autoref{tab:ramsdistancesresults}, we report \F{} based on distance on the development set and find that performance on distant arguments is comparable to performance on local arguments, demonstrating the model's ability to handle non-local arguments.

\begin{figure}[t]
\centering
\includegraphics[width=0.85\linewidth]{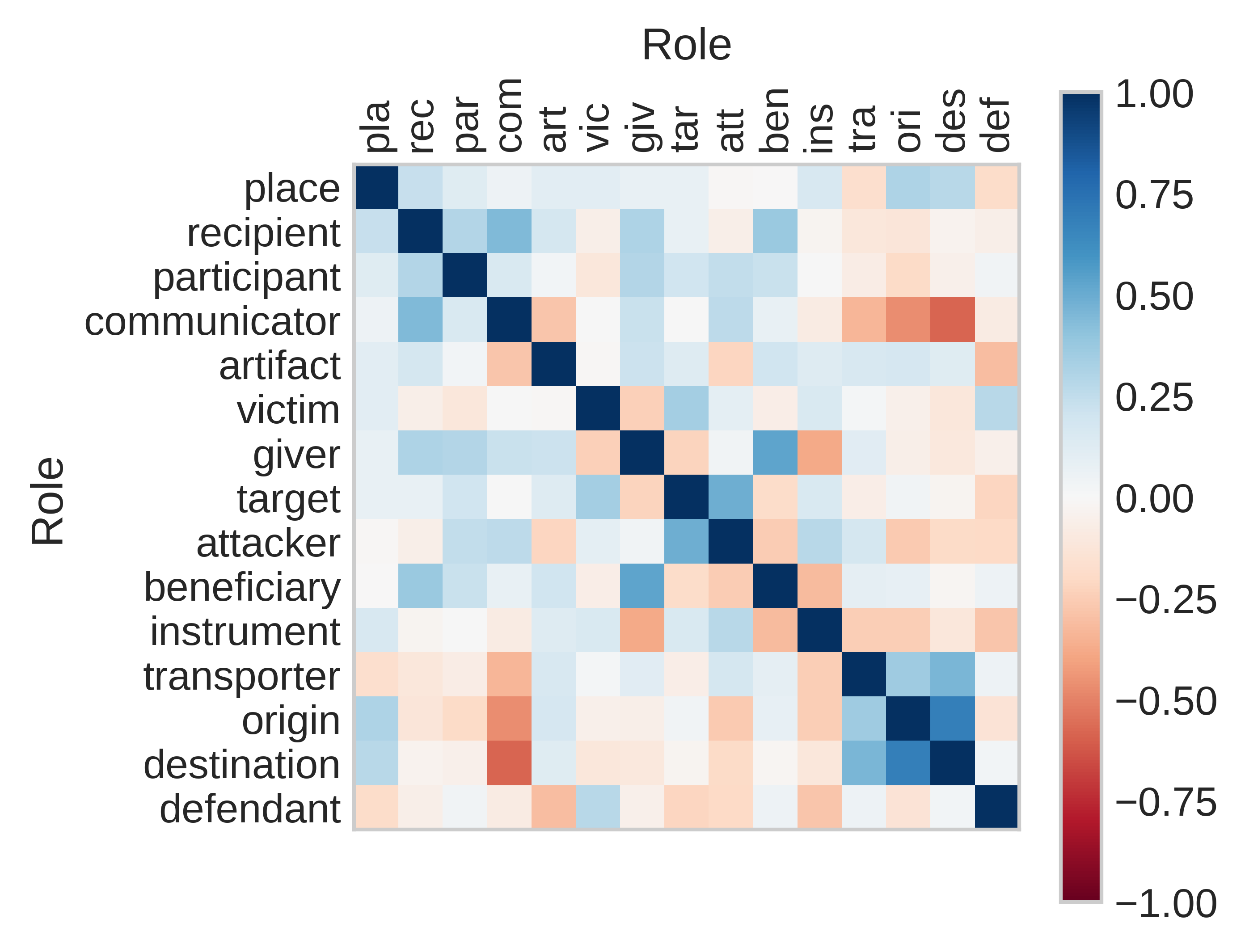}
\includegraphics[width=0.85\linewidth]{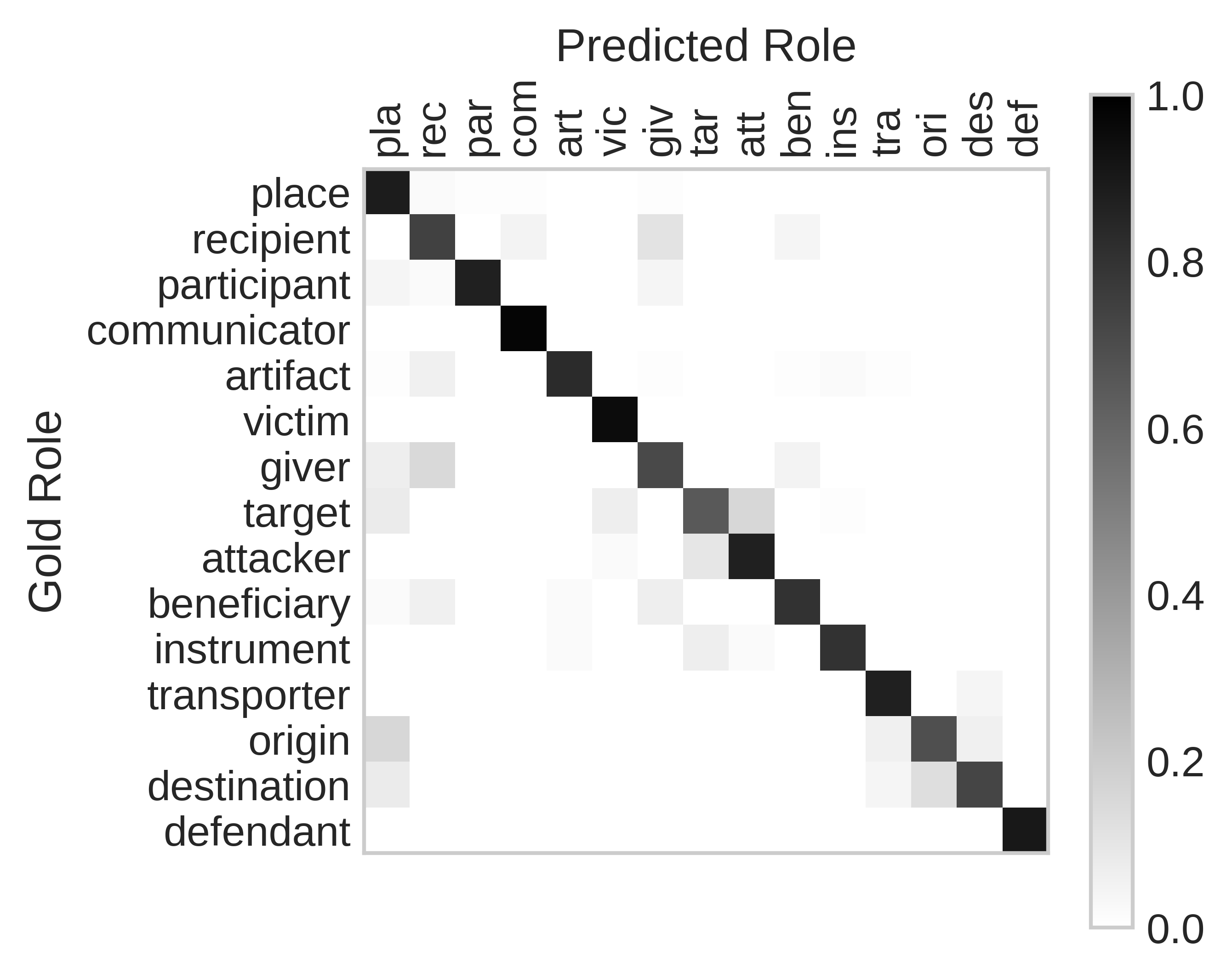}
\caption{Embedding similarity (top) and row-normalized confusion (bottom) between roles for the 15 most frequent roles with our model. The full figures are included in \autoref{appendix:matrices}. Best viewed in color.}
\label{fig:embsim}
\end{figure}

\paragraph{Role Embeddings and Confusion} We present in \autoref{fig:embsim} the cosine similarities between the learned 50-dimensional role embeddings in our model and also the errors made by the model under argmax decoding on the dev set.\footnote{Analysis of the confusion matrix with type-constrained decoding is less meaningful because the constraints, which rely on gold event types, filter out major classes of errors.} Some roles are highly correlated. For example, \ont{origin} and \ont{destination} have the most similar embeddings, possibly because they co-occur frequently and have the same entity type. Conversely, negatively correlated roles have different entity types or occur in different events, such as \ont{communicator} compared to \ont{destination} and \ont{artifact}. We also observe that incorrect predictions are made more often between highly correlated roles and err on the side of the more frequent role, as most errors occur below the diagonal.

\begin{table}[t]
\small
\centering
\begin{tabular}{p{7.25cm}}
\toprule
The $\role[]{EU's leaders}{participant}$ in Brussels are expected to play hardball in negotiating Britain's exit, to send a message to other states that might be contemplating a similar move. ``Informal meeting of EU 27 next week without PM in the room to decide common \underline{\textbf{\textit{negotiating}}} position vs $\role[]{UK}{participant}$ on exit negotiations'' ---Faisal Islam.\\  
\midrule
SPEAKER: I'm Mary Ann Mendoza, the mother of $\role[]{Sergeant Brandon Mendoza}{driverpassenger}$, who was killed in a violent head-on \underline{\textbf{\textit{collision}}} in $\role[]{Mesa}{place}$.\\  
\bottomrule
\end{tabular}
\caption{Two examples of correct predictions on the development set.}
\label{tab:ramsexamples}
\end{table}
\begin{table}[t]
\small
\centering
\begin{tabular}{p{7.25cm}}
\toprule
``Many people are saying that the $\role[]{Iranians}{killer}$ killed the scientist who helped the US because of Hillary Clinton's hacked emails.'' ---8 August, Twitter. $\role[]{Shahran Amiri}{victim, defendant}$, the nuclear scientist \underline{\textbf{\textit{executed}}} in $\role[]{Iran}{place}$ last week, ... \\ 
\midrule
``Many people are saying that the $\role[]{Iranians}{judgecourt}$ killed the scientist who $\role[]{helped the US}{crime}$ because of Hillary Clinton's hacked emails.'' ---8 August, Twitter. $\role[]{Shahran Amiri}{defendant}$, the nuclear scientist \underline{\textbf{\textit{executed}}} in $\role[]{Iran}{place}$ last week, ... \\
\bottomrule
\end{tabular}
\caption{A partially correct prediction (top) and its corresponding gold annotations (bottom).}
\label{tab:ramserrors}
\end{table}
\paragraph{Examples} We present predictions from the development set which demonstrate some phenomena of interest. These are made without TCD, illustrating the model's predictions without knowledge of gold event types.

In \autoref{tab:ramsexamples}, the first example demonstrates the model's ability to link a non-local argument which occurs in the sentence before the trigger. Greedy decoding helps the model find multiple arguments satisfying the same \ont{participant} role, which also appear on either side of the trigger. In the second example, the model correctly predicts the \ont{driverpassenger}, one of the rarer roles in \datasetname{} (17 instances in the training set), consistent with the gold \ont{AccidentCrash} event type.

In \autoref{tab:ramserrors}, the model fills roles corresponding to both the \ont{Death} and the gold \ont{JudicialConsequences} event types, thereby mixing roles from different event types. The predictions are plausible when interpreted in context and would be more accurate under TCD.

\subsection{AIDA Phase~1}
\label{sec:rams:aida}
We also investigate how well \datasetname{} serves as pre-training data for AIDA-1. A model using the hyperparameters of our best-performing \datasetname{} model and trained on just English AIDA-1 Practice data achieves 19.1~\F{} on the English AIDA-1 Eval data under greedy decoding and 18.2~\F{} with TCD. When our best-performing \datasetname{} model is fine-tuned to the AIDA task by further training on the AIDA-1 data, performance is improved to 24.4~\F{} under greedy decoding and 24.8~\F{} with TCD. The crowdsourced annotations in \datasetname{} are therefore of sufficient quality to serve as augmentation to LDC's AIDA-1. Experimental details are available in \autoref{appendix:aida}.

\section{Other Datasets}

\subsection{Beyond NomBank}

The Beyond NomBank (BNB) dataset collected by~\citet{gerber-chai-2010-beyond} and refined by~\citet{gerber-chai-2012-semantic} contains nominal predicates (event triggers) and multi-sentence arguments, both of which are properties shared with \datasetname{}.

To accommodate our formulation of the argument linking task, we modify the BNB data in two ways: 1) we merge ``split'' arguments, which in all but one case are already contiguous spans; and 2) we reduce each cluster of acceptable argument fillers to a set containing only the argument closest to the trigger. We also make modifications to the data splits for purposes of evaluation. \citet{gerber-chai-2012-semantic} suggest evaluation be done using cross-validation on shuffled data, but this may cause document information to leak between the train and evaluation folds. To prevent such leakage and to have a development set for hyperparameter tuning, we separate the data into train, dev, and test splits with no document overlap. Additional data processing details and hyperparameters are given in~\autoref{appendix:bnb}. When given gold triggers and argument spans, our model achieves 75.4~\F{} on dev data and 76.6~\F{} on test data.

\subsection{Gun Violence Database}
\label{sec:gvdb}

The Gun Violence Database (GVDB) \cite{pavlick-etal-2016-gun} is a collection of news articles from the early 2000s to 2016 with annotations specifically related to a \emph{gun violence} event. We split the corpus chronologically into a training set of 5,056 articles, a development set of 400, and a test set of 500. We use this dataset to perform a MUC-style information extraction task \cite{sundheim-1992-overview}. While GVDB's schema permits any number of shooters or victims, we simply predict the first mention of each type. \citet{pavlick-etal-2016-gun} perform evaluation in two settings: a \emph{strict} match is awarded if the predicted string matches the gold string exactly, while an \emph{approximate} match is awarded if either string contains the other.

Assuming each document contains a single gun violence event triggered by the full document, our goal is to predict the value (argument) for each slot (role) for the event. As each slot is filled by exactly one value, we use argmax decoding.

\begin{table}[t]
    \centering
    \small
    \begin{tabular}{lcc}
    \toprule
    {Field} & {Baseline*} & {Our Model}  \\
    \midrule
    Victim Name & 9.3 (54.1) & 62.2 (69.6)\\
    Shooter Name & 4.7 (24.1) & 53.1 (57.8)\\
    Location &  12.2 (18.9) & 34.9 (63.3) \\
    Time & 68.1 (69.3) & 62.9 (69.4) \\
    Weapon & 1.1 (17.9) & 32.5 (49.6)\\
    \bottomrule
    \end{tabular}
    \caption{Strict (and approximate) match \F{} on GVDB. Due to the different data splits and evaluation conditions, we are not directly comparable to the baseline~\cite{pavlick-etal-2016-gun}, provided only for reference.}
    \label{tab:gvdb_results_short}
\end{table}

While the baseline experiments of~\citet{pavlick-etal-2016-gun} made sentence-level predictions focusing on five attributes, we make document-level predictions and consider the larger set of attributes. \autoref{tab:gvdb_results_short} shows our model's performance on the shared subset of attributes, but the numerical values are not directly comparable because the prior work makes predictions on the full dataset and also combines some roles. Our results show that our model is suitable for information extraction tasks like slot filling. \autoref{appendix:gvdb} contains information on hyperparameters and performance on the full set of roles. To our knowledge, our results are a substantial improvement over prior attempts to predict attributes of gun violence event reports, and we make our models available in the hopes of assisting social scientists in their corpus studies.

\section{Conclusion}
We introduced a novel model for document-level argument linking. Because of the small amount of existing data for the task, to support training our neural framework we constructed the \datasetname{} dataset consisting of 9,124 events covering 139 event types. Our model outperforms strong baselines on \datasetname{}, and we also illustrated its applicability to a variety of related datasets. We hope that \datasetname{} will stimulate further work on multi-sentence argument linking.

\section*{Acknowledgments}
We thank Craig Harman for his help in developing the annotation interface. We also thank Tongfei Chen, Yunmo Chen, members of JHU CLSP, and the anonymous reviewers for their helpful discussions and feedback. This work was supported in part by DARPA AIDA (FA8750-18-2-0015) and IARPA BETTER (\#2019-19051600005). The views and conclusions contained in this work are those of the authors and should not be interpreted as necessarily representing the official policies, either expressed or implied, or endorsements of DARPA, ODNI, IARPA, or the U.S. Government. The U.S. Government is authorized to reproduce and distribute reprints for governmental purposes notwithstanding any copyright annotation therein.

\bibliography{acl2020}

\begin{thebibliography}{40}
\expandafter\ifx\csname natexlab\endcsname\relax\def\natexlab#1{#1}\fi

\bibitem[{Baker et~al.(1998)Baker, Fillmore, and
  Lowe}]{baker-etal-1998-berkeley-framenet}
Collin~F. Baker, Charles~J. Fillmore, and John~B. Lowe. 1998.
\newblock \href {https://www.aclweb.org/anthology/C98-1013} {The {B}erkeley
  {F}rame{N}et project}.
\newblock In \emph{{COLING} 1998 Volume 1: The 17th International Conference on
  Computational Linguistics}.

\bibitem[{Burchardt et~al.(2006)Burchardt, Erk, Frank, Kowalski, and
  Pado}]{burchardt-etal-2006-salto}
Aljoscha Burchardt, Katrin Erk, Anette Frank, Andrea Kowalski, and Sebastian
  Pado. 2006.
\newblock \href {http://www.lrec-conf.org/proceedings/lrec2006/pdf/341_pdf.pdf}
  {{SALTO} - a versatile multi-level annotation tool}.
\newblock In \emph{Proceedings of the Fifth International Conference on
  Language Resources and Evaluation ({LREC}{'}06)}, Genoa, Italy. European
  Language Resources Association (ELRA).

\bibitem[{Chen et~al.(2015)Chen, Xu, Liu, Zeng, and
  Zhao}]{chen-etal-2015-event}
Yubo Chen, Liheng Xu, Kang Liu, Daojian Zeng, and Jun Zhao. 2015.
\newblock \href {https://doi.org/10.3115/v1/P15-1017} {Event extraction via
  dynamic multi-pooling convolutional neural networks}.
\newblock In \emph{Proceedings of the 53rd Annual Meeting of the Association
  for Computational Linguistics and the 7th International Joint Conference on
  Natural Language Processing (Volume 1: Long Papers)}, pages 167--176,
  Beijing, China. Association for Computational Linguistics.

\bibitem[{Cheng and Erk(2018)}]{cheng-erk-2018-implicit}
Pengxiang Cheng and Katrin Erk. 2018.
\newblock \href {https://doi.org/10.18653/v1/N18-1076} {Implicit argument
  prediction with event knowledge}.
\newblock In \emph{Proceedings of the 2018 Conference of the North {A}merican
  Chapter of the Association for Computational Linguistics: Human Language
  Technologies, Volume 1 (Long Papers)}, pages 831--840, New Orleans,
  Louisiana. Association for Computational Linguistics.

\bibitem[{Cheng and Erk(2019)}]{cheng2019implicit}
Pengxiang Cheng and Katrin Erk. 2019.
\newblock Implicit argument prediction as reading comprehension.
\newblock In \emph{Proceedings of the AAAI Conference on Artificial
  Intelligence}, volume~33, pages 6284--6291.

\bibitem[{Das et~al.(2010)Das, Schneider, Chen, and
  Smith}]{das-etal-2010-probabilistic}
Dipanjan Das, Nathan Schneider, Desai Chen, and Noah~A. Smith. 2010.
\newblock \href {https://www.aclweb.org/anthology/N10-1138} {Probabilistic
  frame-semantic parsing}.
\newblock In \emph{Human Language Technologies: The 2010 Annual Conference of
  the North {A}merican Chapter of the Association for Computational
  Linguistics}, pages 948--956, Los Angeles, California. Association for
  Computational Linguistics.

\bibitem[{Devlin et~al.(2018)Devlin, Chang, Lee, and Toutanova}]{Devlin18BERT}
Jacob Devlin, Ming{-}Wei Chang, Kenton Lee, and Kristina Toutanova. 2018.
\newblock \href {http://arxiv.org/abs/1810.04805} {{BERT:} pre-training of deep
  bidirectional transformers for language understanding}.
\newblock \emph{CoRR}, abs/1810.04805.

\bibitem[{Feizabadi and Pad{\'o}(2014)}]{feizabadi-pado-2014-crowdsourcing}
Parvin~Sadat Feizabadi and Sebastian Pad{\'o}. 2014.
\newblock \href {https://doi.org/10.3115/v1/E14-4044} {Crowdsourcing annotation
  of non-local semantic roles}.
\newblock In \emph{Proceedings of the 14th Conference of the {E}uropean Chapter
  of the Association for Computational Linguistics, volume 2: Short Papers},
  pages 226--230, Gothenburg, Sweden. Association for Computational
  Linguistics.

\bibitem[{Feizabadi and Pad{\'o}(2015)}]{feizabadi-pado-2015-combining}
Parvin~Sadat Feizabadi and Sebastian Pad{\'o}. 2015.
\newblock \href {https://doi.org/10.18653/v1/S15-1005} {Combining seemingly
  incompatible corpora for implicit semantic role labeling}.
\newblock In \emph{Proceedings of the Fourth Joint Conference on Lexical and
  Computational Semantics}, pages 40--50, Denver, Colorado. Association for
  Computational Linguistics.

\bibitem[{Fillmore(1986)}]{fillmore1986pragmatically}
Charles~J Fillmore. 1986.
\newblock Pragmatically controlled zero anaphora.
\newblock In \emph{Annual Meeting of the Berkeley Linguistics Society},
  volume~12, pages 95--107.

\bibitem[{Fillmore et~al.(2002)Fillmore, Baker, and
  Sato}]{fillmore-etal-2002-framenet}
Charles~J. Fillmore, Collin~F. Baker, and Hiroaki Sato. 2002.
\newblock \href {http://www.lrec-conf.org/proceedings/lrec2002/pdf/140.pdf}
  {The {F}rame{N}et database and software tools}.
\newblock In \emph{Proceedings of the Third International Conference on
  Language Resources and Evaluation ({LREC}{'}02)}, Las Palmas, Canary Islands
  - Spain. European Language Resources Association (ELRA).

\bibitem[{Gerber and Chai(2010)}]{gerber-chai-2010-beyond}
Matthew Gerber and Joyce Chai. 2010.
\newblock \href {https://www.aclweb.org/anthology/P10-1160} {Beyond
  {N}om{B}ank: A study of implicit arguments for nominal predicates}.
\newblock In \emph{Proceedings of the 48th Annual Meeting of the Association
  for Computational Linguistics}, pages 1583--1592, Uppsala, Sweden.
  Association for Computational Linguistics.

\bibitem[{Gerber and Chai(2012)}]{gerber-chai-2012-semantic}
Matthew Gerber and Joyce~Y. Chai. 2012.
\newblock \href {https://doi.org/10.1162/COLI_a_00110} {Semantic role labeling
  of implicit arguments for nominal predicates}.
\newblock \emph{Computational Linguistics}, 38(4):755--798.

\bibitem[{He et~al.(2018)He, Lee, Levy, and Zettlemoyer}]{he-etal-2018-jointly}
Luheng He, Kenton Lee, Omer Levy, and Luke Zettlemoyer. 2018.
\newblock \href {https://www.aclweb.org/anthology/P18-2058} {Jointly predicting
  predicates and arguments in neural semantic role labeling}.
\newblock In \emph{Proceedings of the 56th Annual Meeting of the Association
  for Computational Linguistics (Volume 2: Short Papers)}, pages 364--369,
  Melbourne, Australia. Association for Computational Linguistics.

\bibitem[{Kingma and Ba(2015)}]{DBLP:journals/corr/KingmaB14}
Diederik~P. Kingma and Jimmy Ba. 2015.
\newblock \href {http://arxiv.org/abs/1412.6980} {Adam: {A} method for
  stochastic optimization}.
\newblock In \emph{3rd International Conference on Learning Representations,
  {ICLR} 2015, San Diego, CA, USA, May 7-9, 2015, Conference Track
  Proceedings}.

\bibitem[{Kitaev and Klein(2018)}]{kitaev-klein-2018-constituency}
Nikita Kitaev and Dan Klein. 2018.
\newblock \href {https://doi.org/10.18653/v1/P18-1249} {Constituency parsing
  with a self-attentive encoder}.
\newblock In \emph{Proceedings of the 56th Annual Meeting of the Association
  for Computational Linguistics (Volume 1: Long Papers)}, pages 2676--2686,
  Melbourne, Australia. Association for Computational Linguistics.

\bibitem[{Knight et~al.(2020)Knight, Badarau, Baranescu, Bonial, Bardocz,
  Griffitt, Hermjakob, Marcu, Palmer, O'{G}orman, and
  Schneider}]{knight2020amr}
Kevin Knight, Bianca Badarau, Laura Baranescu, Claire Bonial, Madalina Bardocz,
  Kira Griffitt, Ulf Hermjakob, Daniel Marcu, Martha Palmer, Tim O'{G}orman,
  and Nathan Schneider. 2020.
\newblock {A}bstract {M}eaning {R}epresentation ({AMR}) annotation release 3.0
  {LDC2020T02}.
\newblock \emph{Linguistic Data Consortium, Philadelphia, PA}.

\bibitem[{Lee et~al.(2017)Lee, He, Lewis, and Zettlemoyer}]{lee-etal-2017-end}
Kenton Lee, Luheng He, Mike Lewis, and Luke Zettlemoyer. 2017.
\newblock \href {https://doi.org/10.18653/v1/D17-1018} {End-to-end neural
  coreference resolution}.
\newblock In \emph{Proceedings of the 2017 Conference on Empirical Methods in
  Natural Language Processing}, pages 188--197, Copenhagen, Denmark.
  Association for Computational Linguistics.

\bibitem[{Lee et~al.(2018)Lee, He, and Zettlemoyer}]{lee-etal-2018-higher}
Kenton Lee, Luheng He, and Luke Zettlemoyer. 2018.
\newblock \href {https://doi.org/10.18653/v1/N18-2108} {Higher-order
  coreference resolution with coarse-to-fine inference}.
\newblock In \emph{Proceedings of the 2018 Conference of the North {A}merican
  Chapter of the Association for Computational Linguistics: Human Language
  Technologies, Volume 2 (Short Papers)}, pages 687--692, New Orleans,
  Louisiana. Association for Computational Linguistics.

\bibitem[{Nguyen et~al.(2016)Nguyen, Cho, and
  Grishman}]{nguyen-etal-2016-joint-event}
Thien~Huu Nguyen, Kyunghyun Cho, and Ralph Grishman. 2016.
\newblock \href {https://doi.org/10.18653/v1/N16-1034} {Joint event extraction
  via recurrent neural networks}.
\newblock In \emph{Proceedings of the 2016 Conference of the North {A}merican
  Chapter of the Association for Computational Linguistics: Human Language
  Technologies}, pages 300--309, San Diego, California. Association for
  Computational Linguistics.

\bibitem[{O{'}Gorman et~al.(2018)O{'}Gorman, Regan, Griffitt, Hermjakob,
  Knight, and Palmer}]{ogorman-etal-2018-amr}
Tim O{'}Gorman, Michael Regan, Kira Griffitt, Ulf Hermjakob, Kevin Knight, and
  Martha Palmer. 2018.
\newblock \href {https://www.aclweb.org/anthology/C18-1313} {{AMR} beyond the
  sentence: the multi-sentence {AMR} corpus}.
\newblock In \emph{Proceedings of the 27th International Conference on
  Computational Linguistics}, pages 3693--3702, Santa Fe, New Mexico, USA.
  Association for Computational Linguistics.

\bibitem[{O'Gorman(2019)}]{OGorman19}
Timothy~J O'Gorman. 2019.
\newblock \emph{Bringing Together Computational and Linguistic Models of
  Implicit Role Interpretation}.
\newblock {PhD} dissertation, University of Colorado at Boulder.

\bibitem[{Ouchi et~al.(2018)Ouchi, Shindo, and
  Matsumoto}]{ouchi-etal-2018-span}
Hiroki Ouchi, Hiroyuki Shindo, and Yuji Matsumoto. 2018.
\newblock \href {https://doi.org/10.18653/v1/D18-1191} {A span selection model
  for semantic role labeling}.
\newblock In \emph{Proceedings of the 2018 Conference on Empirical Methods in
  Natural Language Processing}, pages 1630--1642, Brussels, Belgium.
  Association for Computational Linguistics.

\bibitem[{Palmer et~al.(2005)Palmer, Gildea, and
  Kingsbury}]{palmer-etal-2005-proposition}
Martha Palmer, Daniel Gildea, and Paul Kingsbury. 2005.
\newblock \href {https://doi.org/10.1162/0891201053630264} {The proposition
  bank: An annotated corpus of semantic roles}.
\newblock \emph{Computational Linguistics}, 31(1):71--106.

\bibitem[{Pavlick et~al.(2016)Pavlick, Ji, Pan, and
  Callison-Burch}]{pavlick-etal-2016-gun}
Ellie Pavlick, Heng Ji, Xiaoman Pan, and Chris Callison-Burch. 2016.
\newblock \href {https://doi.org/10.18653/v1/D16-1106} {The gun violence
  database: A new task and data set for {NLP}}.
\newblock In \emph{Proceedings of the 2016 Conference on Empirical Methods in
  Natural Language Processing}, pages 1018--1024, Austin, Texas. Association
  for Computational Linguistics.

\bibitem[{Pennington et~al.(2014)Pennington, Socher, and
  Manning}]{pennington-etal-2014-glove}
Jeffrey Pennington, Richard Socher, and Christopher Manning. 2014.
\newblock \href {https://doi.org/10.3115/v1/D14-1162} {{G}love: Global vectors
  for word representation}.
\newblock In \emph{Proceedings of the 2014 Conference on Empirical Methods in
  Natural Language Processing ({EMNLP})}, pages 1532--1543, Doha, Qatar.
  Association for Computational Linguistics.

\bibitem[{Peters et~al.(2018)Peters, Neumann, Iyyer, Gardner, Clark, Lee, and
  Zettlemoyer}]{peters-etal-2018-deep}
Matthew Peters, Mark Neumann, Mohit Iyyer, Matt Gardner, Christopher Clark,
  Kenton Lee, and Luke Zettlemoyer. 2018.
\newblock \href {https://doi.org/10.18653/v1/N18-1202} {Deep contextualized
  word representations}.
\newblock In \emph{Proceedings of the 2018 Conference of the North {A}merican
  Chapter of the Association for Computational Linguistics: Human Language
  Technologies, Volume 1 (Long Papers)}, pages 2227--2237, New Orleans,
  Louisiana. Association for Computational Linguistics.

\bibitem[{Pradhan et~al.(2013)Pradhan, Moschitti, Xue, Ng, Bj{\"o}rkelund,
  Uryupina, Zhang, and Zhong}]{pradhan-etal-2013-towards}
Sameer Pradhan, Alessandro Moschitti, Nianwen Xue, Hwee~Tou Ng, Anders
  Bj{\"o}rkelund, Olga Uryupina, Yuchen Zhang, and Zhi Zhong. 2013.
\newblock \href {https://www.aclweb.org/anthology/W13-3516} {Towards robust
  linguistic analysis using {O}nto{N}otes}.
\newblock In \emph{Proceedings of the Seventeenth Conference on Computational
  Natural Language Learning}, pages 143--152, Sofia, Bulgaria. Association for
  Computational Linguistics.

\bibitem[{Radford et~al.(2019)Radford, Wu, Child, Luan, Amodei, and
  Sutskever}]{radford2019language}
Alec Radford, Jeff Wu, Rewon Child, David Luan, Dario Amodei, and Ilya
  Sutskever. 2019.
\newblock Language models are unsupervised multitask learners.

\bibitem[{Roth and Frank(2013)}]{roth-frank-2013-automatically}
Michael Roth and Anette Frank. 2013.
\newblock \href {https://www.aclweb.org/anthology/S13-1043} {Automatically
  identifying implicit arguments to improve argument linking and coherence
  modeling}.
\newblock In \emph{Second Joint Conference on Lexical and Computational
  Semantics (*{SEM}), Volume 1: Proceedings of the Main Conference and the
  Shared Task: Semantic Textual Similarity}, pages 306--316, Atlanta, Georgia,
  USA. Association for Computational Linguistics.

\bibitem[{Ruppenhofer et~al.(2010)Ruppenhofer, Sporleder, Morante, Baker, and
  Palmer}]{ruppenhofer-etal-2010-semeval}
Josef Ruppenhofer, Caroline Sporleder, Roser Morante, Collin Baker, and Martha
  Palmer. 2010.
\newblock \href {https://www.aclweb.org/anthology/S10-1008} {{S}em{E}val-2010
  task 10: Linking events and their participants in discourse}.
\newblock In \emph{Proceedings of the 5th International Workshop on Semantic
  Evaluation}, pages 45--50, Uppsala, Sweden. Association for Computational
  Linguistics.

\bibitem[{Schenk and Chiarcos(2016)}]{schenk-chiarcos-2016-unsupervised}
Niko Schenk and Christian Chiarcos. 2016.
\newblock \href {https://doi.org/10.18653/v1/N16-1173} {Unsupervised learning
  of prototypical fillers for implicit semantic role labeling}.
\newblock In \emph{Proceedings of the 2016 Conference of the North {A}merican
  Chapter of the Association for Computational Linguistics: Human Language
  Technologies}, pages 1473--1479, San Diego, California. Association for
  Computational Linguistics.

\bibitem[{Silberer and Frank(2012)}]{silberer-frank-2012-casting}
Carina Silberer and Anette Frank. 2012.
\newblock \href {https://www.aclweb.org/anthology/S12-1001} {Casting implicit
  role linking as an anaphora resolution task}.
\newblock In \emph{*{SEM} 2012: The First Joint Conference on Lexical and
  Computational Semantics {--} Volume 1: Proceedings of the main conference and
  the shared task, and Volume 2: Proceedings of the Sixth International
  Workshop on Semantic Evaluation ({S}em{E}val 2012)}, pages 1--10,
  Montr{\'e}al, Canada. Association for Computational Linguistics.

\bibitem[{Song et~al.(2018)Song, Zhang, Wang, and Gildea}]{song-etal-2018-n}
Linfeng Song, Yue Zhang, Zhiguo Wang, and Daniel Gildea. 2018.
\newblock \href {https://doi.org/10.18653/v1/D18-1246} {N-ary relation
  extraction using graph-state {LSTM}}.
\newblock In \emph{Proceedings of the 2018 Conference on Empirical Methods in
  Natural Language Processing}, pages 2226--2235, Brussels, Belgium.
  Association for Computational Linguistics.

\bibitem[{Sundheim(1992)}]{sundheim-1992-overview}
Beth~M. Sundheim. 1992.
\newblock \href {https://www.aclweb.org/anthology/M92-1001} {Overview of the
  fourth message understanding evaluation and conference}.
\newblock In \emph{FOURTH MESSAGE UNDERSTANDING CONFERENCE ({MUC}-4),
  Proceedings of a Conference Held in McLean, Virginia, June 16-18, 1992}.

\bibitem[{Swayamdipta et~al.(2018)Swayamdipta, Thomson, Lee, Zettlemoyer, Dyer,
  and Smith}]{swayamdipta-etal-2018-syntactic}
Swabha Swayamdipta, Sam Thomson, Kenton Lee, Luke Zettlemoyer, Chris Dyer, and
  Noah~A. Smith. 2018.
\newblock \href {https://www.aclweb.org/anthology/D18-1412} {Syntactic
  scaffolds for semantic structures}.
\newblock In \emph{Proceedings of the 2018 Conference on Empirical Methods in
  Natural Language Processing}, pages 3772--3782, Brussels, Belgium.
  Association for Computational Linguistics.

\bibitem[{Tenney et~al.(2019{\natexlab{a}})Tenney, Das, and
  Pavlick}]{tenney-etal-2019-bert}
Ian Tenney, Dipanjan Das, and Ellie Pavlick. 2019{\natexlab{a}}.
\newblock \href {https://doi.org/10.18653/v1/P19-1452} {{BERT} rediscovers the
  classical {NLP} pipeline}.
\newblock In \emph{Proceedings of the 57th Annual Meeting of the Association
  for Computational Linguistics}, pages 4593--4601, Florence, Italy.
  Association for Computational Linguistics.

\bibitem[{Tenney et~al.(2019{\natexlab{b}})Tenney, Xia, Chen, Wang, Poliak,
  McCoy, Kim, Durme, Bowman, Das, and Pavlick}]{tenney2018what}
Ian Tenney, Patrick Xia, Berlin Chen, Alex Wang, Adam Poliak, R~Thomas McCoy,
  Najoung Kim, Benjamin~Van Durme, Sam Bowman, Dipanjan Das, and Ellie Pavlick.
  2019{\natexlab{b}}.
\newblock \href {https://openreview.net/forum?id=SJzSgnRcKX} {What do you learn
  from context? {P}robing for sentence structure in contextualized word
  representations}.
\newblock In \emph{International Conference on Learning Representations}.

\bibitem[{Weischedel et~al.(2013)Weischedel, Palmer, Marcus, Hovy, Pradhan,
  Ramshaw, Xue, Taylor, Kaufman, Franchini, El-{B}achouti, Belvin, and
  Houston}]{weischedel2013ontonotes}
Ralph Weischedel, Martha Palmer, Mitchell Marcus, Eduard Hovy, Sameer Pradhan,
  Lance Ramshaw, Nianwen Xue, Ann Taylor, Jeff Kaufman, Michelle Franchini,
  Mohammed El-{B}achouti, Robert Belvin, and Ann Houston. 2013.
\newblock {OntoNotes} release 5.0 {LDC2013T19}.
\newblock \emph{Linguistic Data Consortium, Philadelphia, PA}.

\bibitem[{Wolfe et~al.(2015)Wolfe, Dredze, and
  Van~Durme}]{wolfe-etal-2015-predicate}
Travis Wolfe, Mark Dredze, and Benjamin Van~Durme. 2015.
\newblock \href {https://doi.org/10.3115/v1/N15-1002} {Predicate argument
  alignment using a global coherence model}.
\newblock In \emph{Proceedings of the 2015 Conference of the North {A}merican
  Chapter of the Association for Computational Linguistics: Human Language
  Technologies}, pages 11--20, Denver, Colorado. Association for Computational
  Linguistics.

\end{thebibliography}
\bibliographystyle{acl_natbib}

\clearpage
\appendix
\section{RAMS Data}
\label{appendix:ramsdata}
\subsection{Collection}
\label{appendix:data:reddit}

On Reddit, users make submissions containing links to news articles, images, videos, or other kinds of documents, and other users may then vote or comment on the submitted content. We collected news articles matching the following criteria: 1) Posted to the \textit{r/politics} sub-forum between January and October 2016; 2) Resulted in threads with at least 25 comments; and 3) Contained at least one mention of the string ``Russia''. The resulting subset of articles tended to describe geopolitical events and relations like the ones in the AIDA ontology. In order to filter out low-quality, fake, or disreputable news articles, we treat the number of comments in the discussion as a signal of information content. Our approach of gathering user-submitted and curated content through Reddit is similar to those used for creating large datasets for language model pre-training \cite{radford2019language}. Documents were split into sentences using \texttt{NLTK 3.4.3}, and sentences were split into tokens using \texttt{SpaCy 2.1.4}.

\subsection{Annotation}
\label{appendix:data:annotation}

\begin{figure*}
    \includegraphics[scale=0.45]{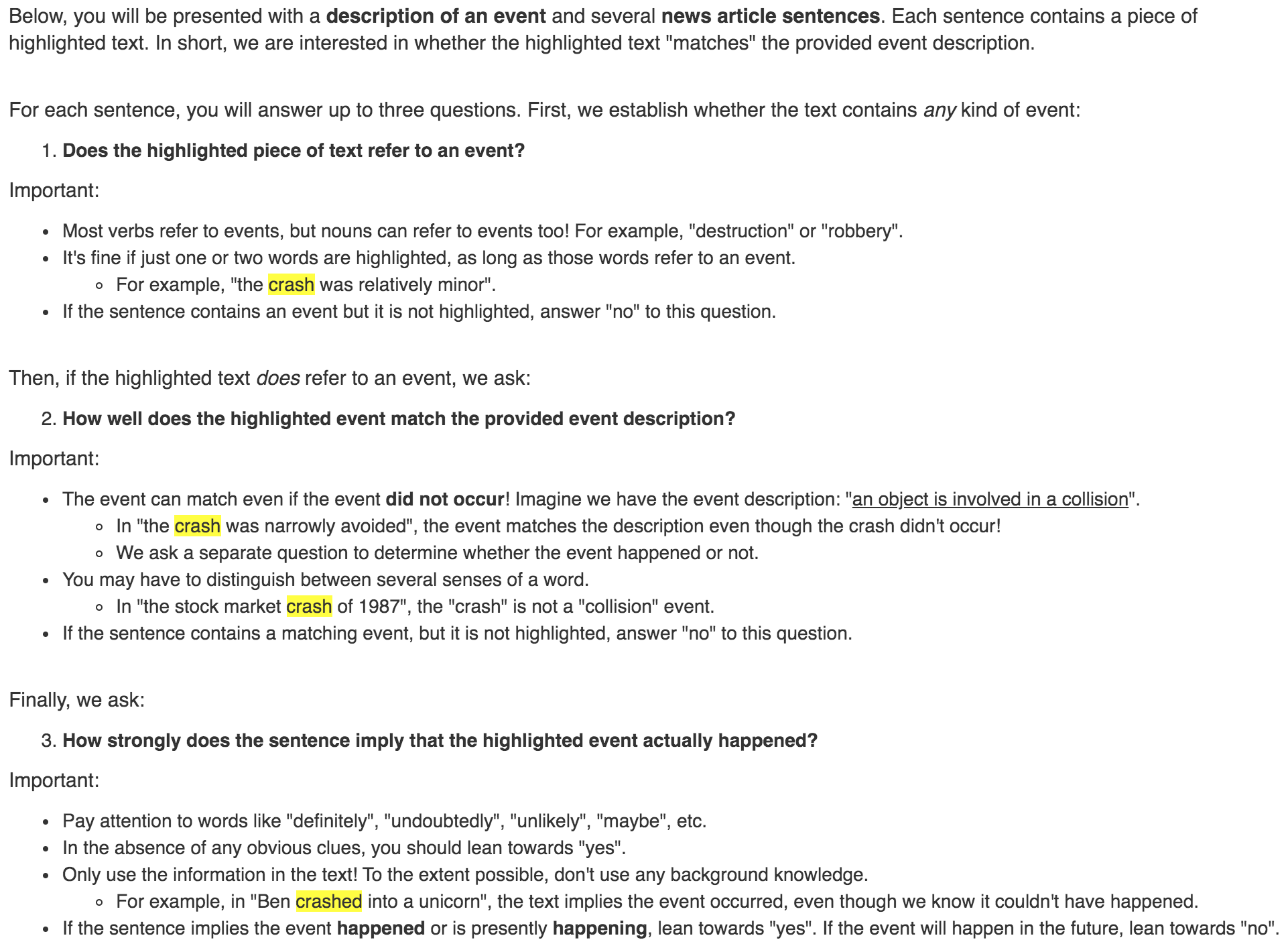}
    \caption{Annotation instructions for determining whether a lexical unit (in context) evokes an event type.}
    \label{fig:ramsannotationeventinstructions}
\end{figure*}

\begin{figure*}
    \includegraphics[scale=0.45]{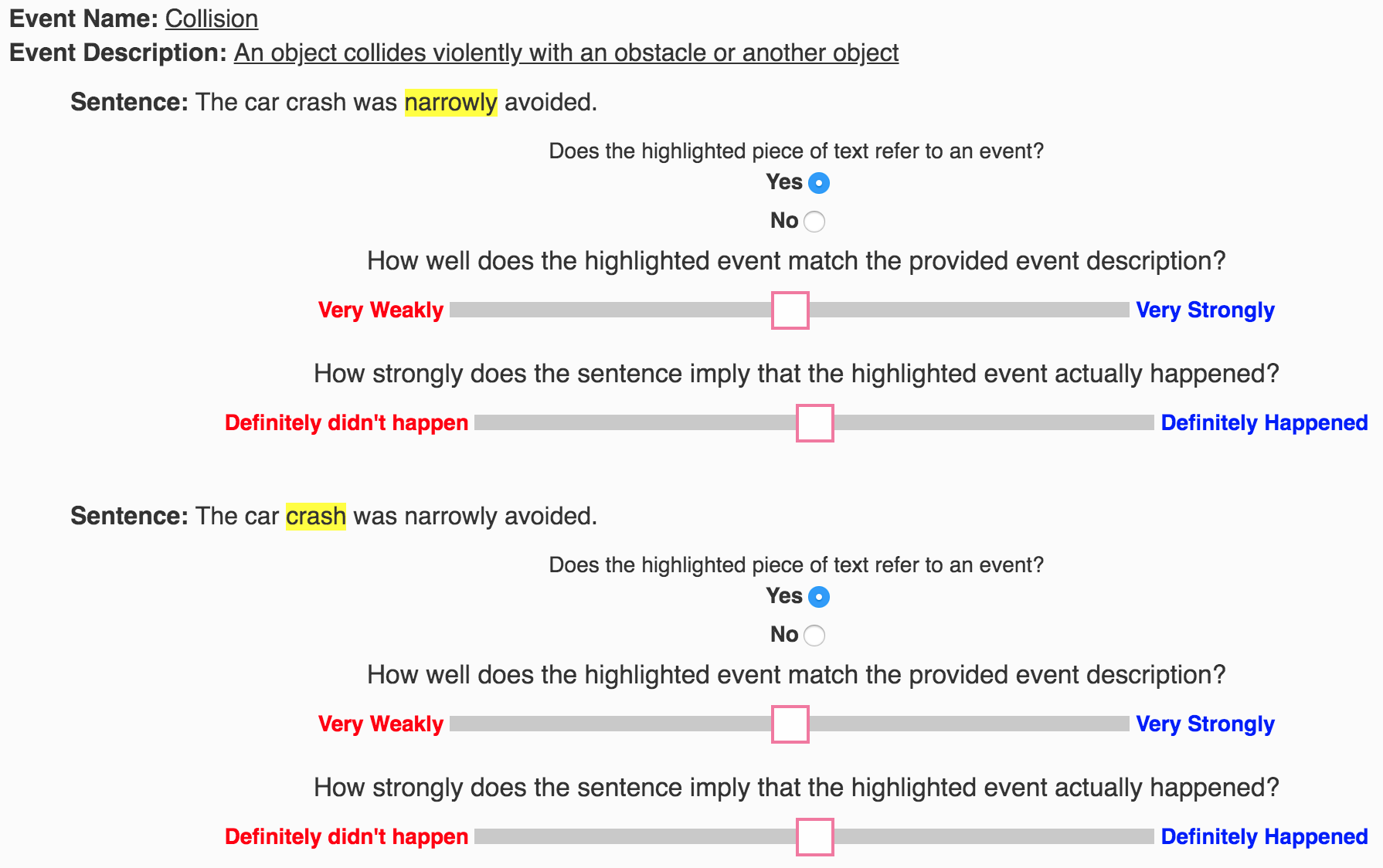}
    \caption{Annotation interface for determining whether a lexical unit (in context) evokes an event type.}
    \label{fig:ramsannotationeventquestions}
\end{figure*}

To assess whether a lexical unit (LU) evoked an event with positive factuality, the vetting task contained an event definition and several candidate sentences, each with a highlighted LU. Annotators were asked to judge how well each highlighted LU, in the context of its sentence, matched the provided event definition. In the same task, they were also asked to assess the factuality of the sentence. Annotation instructions and examples are shown in \autoref{fig:ramsannotationeventinstructions} and \autoref{fig:ramsannotationeventquestions}.

\begin{figure}[h]
\includegraphics[scale=0.34]{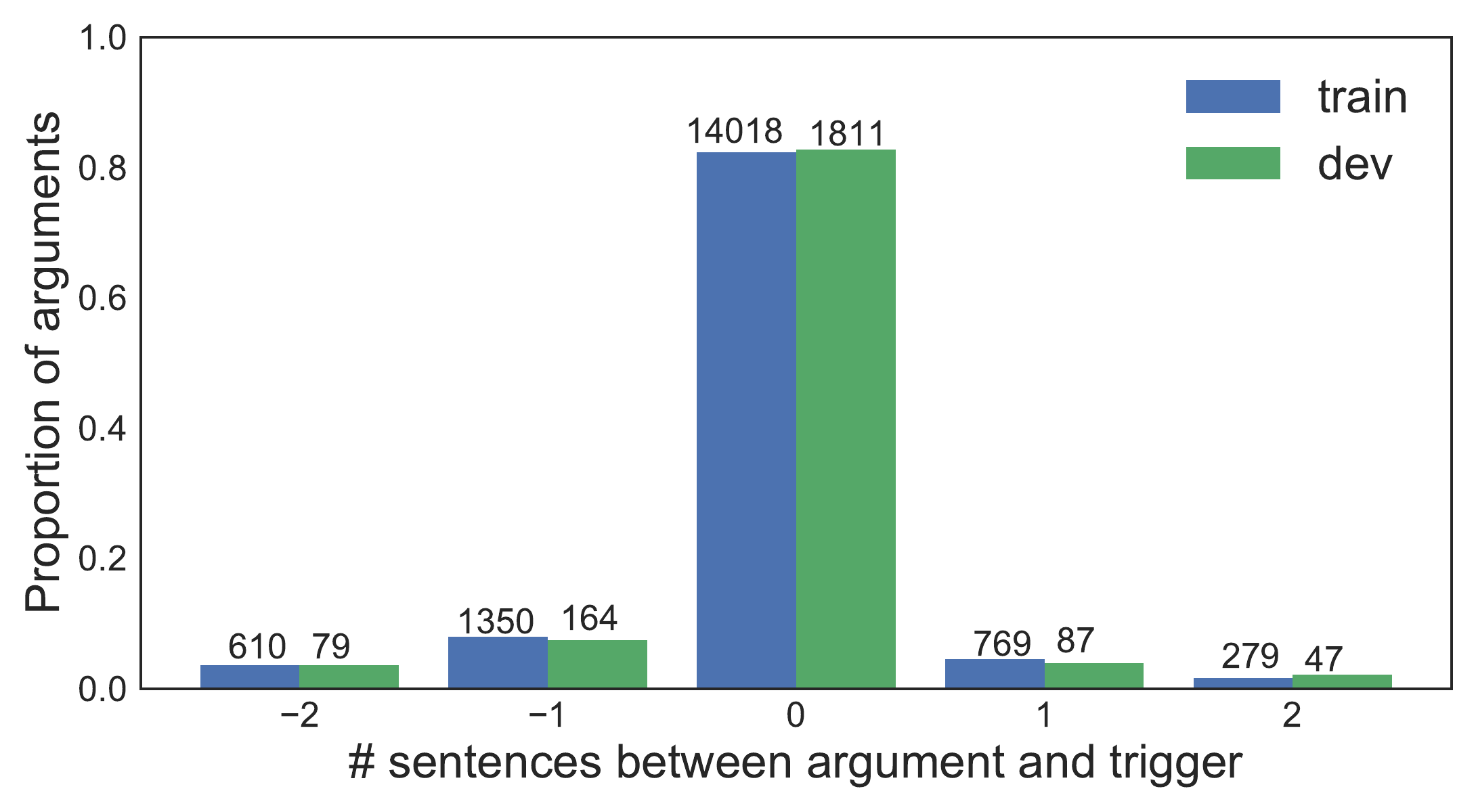}
\caption{Distances between triggers and arguments in \datasetname{} and proportion of arguments at that distance (counts are shown above each bar). Negative distances indicate that the argument occurs before the trigger.}
\label{fig:ramsdistancesstatistics}
\end{figure}

\begin{figure*}
    \includegraphics[scale=0.30]{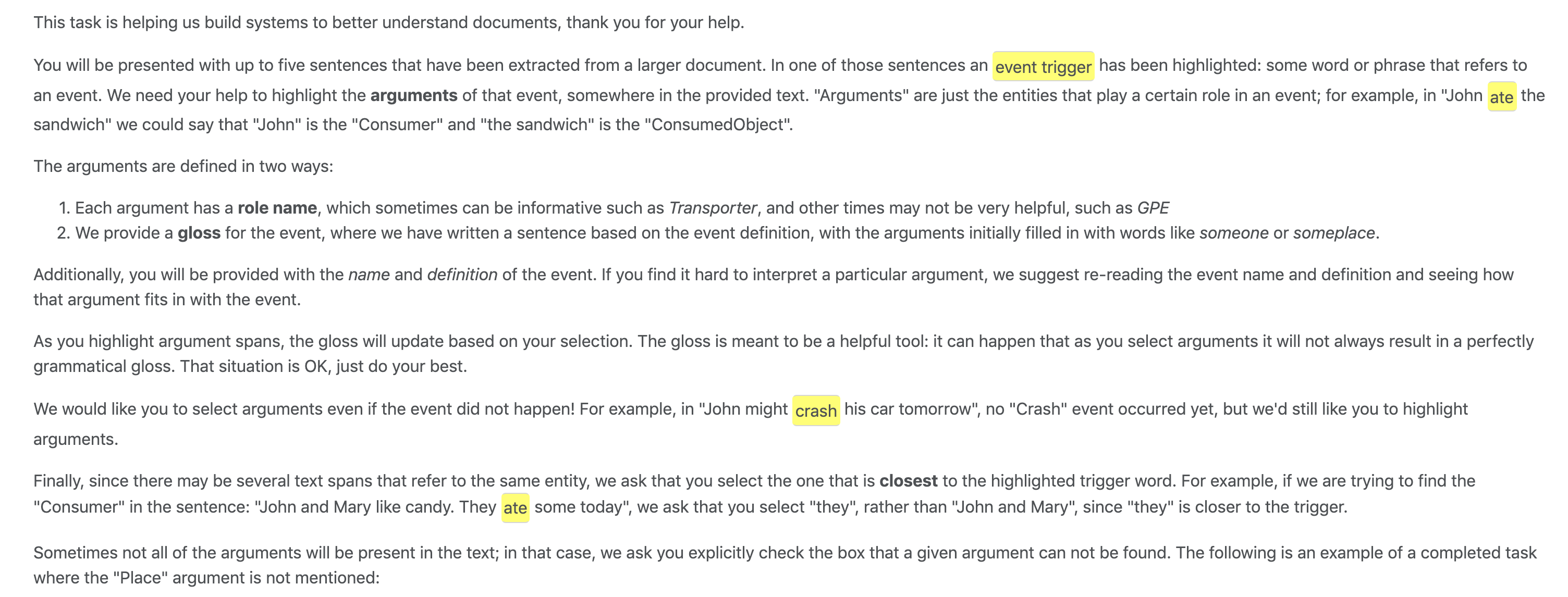}
    \caption{Annotation instructions for selecting arguments for an event.}
    \label{fig:ramsannotationarginstructions}
\end{figure*}

\begin{figure*}
    \includegraphics[scale=0.30]{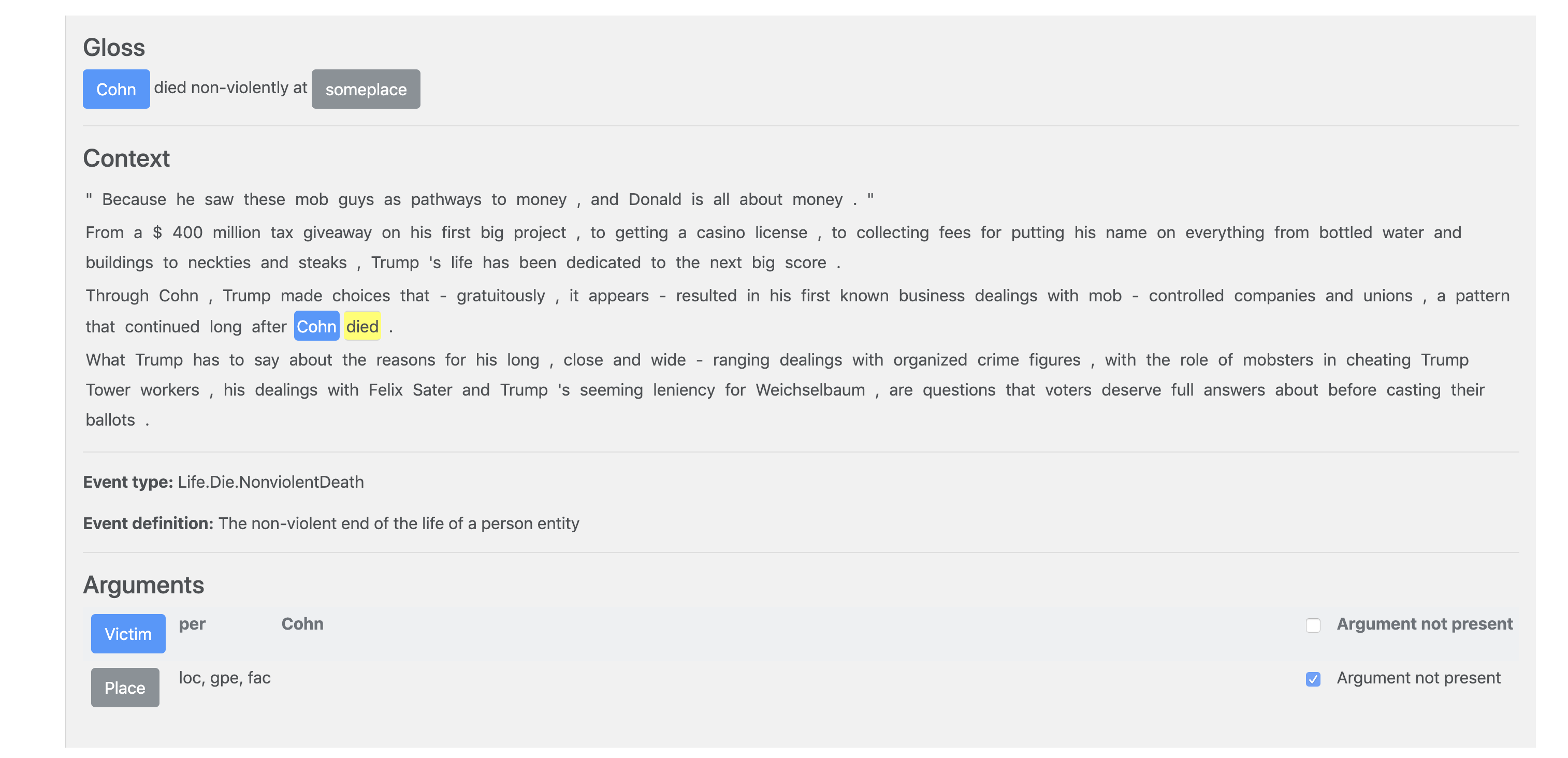}
    \caption{Annotation interface for selecting arguments for an event.}
    \label{fig:ramsannotationargquestions}
\end{figure*}

Each argument selection task contained five tokenized sentences, a contiguous set of tokens marking the trigger, a definition of the event type, and a list of roles and their associated definitions. For each role, annotators were asked whether a corresponding argument was present in the 5-sentence window, and if so, to highlight the argument span that was closest to the event trigger, as there could be multiple. In cases near the beginning or end of a document, annotators were shown up to two sentences before or after the sentence containing the trigger. Annotators were allowed to highlight any set of (within-sentence) contiguous tokens within the 5-sentence window aside from the trigger tokens. The distribution of distances between triggers and arguments is shown in~\autoref{fig:ramsdistancesstatistics}. Annotation instructions and an example are shown in \autoref{fig:ramsannotationarginstructions} and \autoref{fig:ramsannotationargquestions}.

\subsection{Agreement}

We additionally compute the frequency with which annotators agreed a given role was or was not present in the context window. To measure the frequency with which annotators agree whether a given role is present, we treat the majority annotation as the gold standard. Then, we calculated the precision, recall, and \F{} of the annotations.  Across the set of redundantly annotated tasks, there were 83 false negatives, 60 false positives, and 892 true positives, giving a precision of 93.7, recall of 91.5, and an \F{} of 92.6.

\begin{table}[h]
\small
\centering
\begin{tabular}{lcccc}
\toprule
Threshold & Conjunctive & Disjunctive & Start & End \\
\midrule
0 & 55.3 & 78.0 & 59.8 & 73.5 \\
1 & 69.9 & 80.3 & 74.9 & 75.3 \\
2 & 73.9 & 82.0 & 78.2 & 77.8 \\
3 & 76.4 & 83.6 & 80.9 & 79.1  \\
4 & 78.8 & 84.3 & 82.7 & 80.4 \\
\bottomrule
\end{tabular}
\caption{Pairwise span boundary inter-annotator agreement statistics for various span difference thresholds.}
\label{tab:agreement}
\end{table}

We consider a wider range of span difference thresholds, where span difference is calculated by using the absolute difference of the $(start, end)$ token indices from each pair. These are presented in \autoref{tab:agreement}. In conjunctive agreement, both $|start_1 - start_2|$ and $|end_1 - end_2|$ must be less than the given threshold; therefore, conjunctive agreement at threshold 0 is the percent of pairs that exactly agree (55.3\%).  Disjunctive agreement is less strict, requiring that either the absolute difference of start offsets or end offsets must be less than the threshold.  Start and end agreement is determined by considering whether the absolute difference of the pair's start or end offsets (respectively) is within the given threshold.

\subsection{Event and Role Type Coverage}
\label{appendix:data:coverage}

\begin{figure}[t]
\includegraphics[width=0.9\linewidth]{data/type_coverage_freq.pdf}
\includegraphics[width=0.9\linewidth]{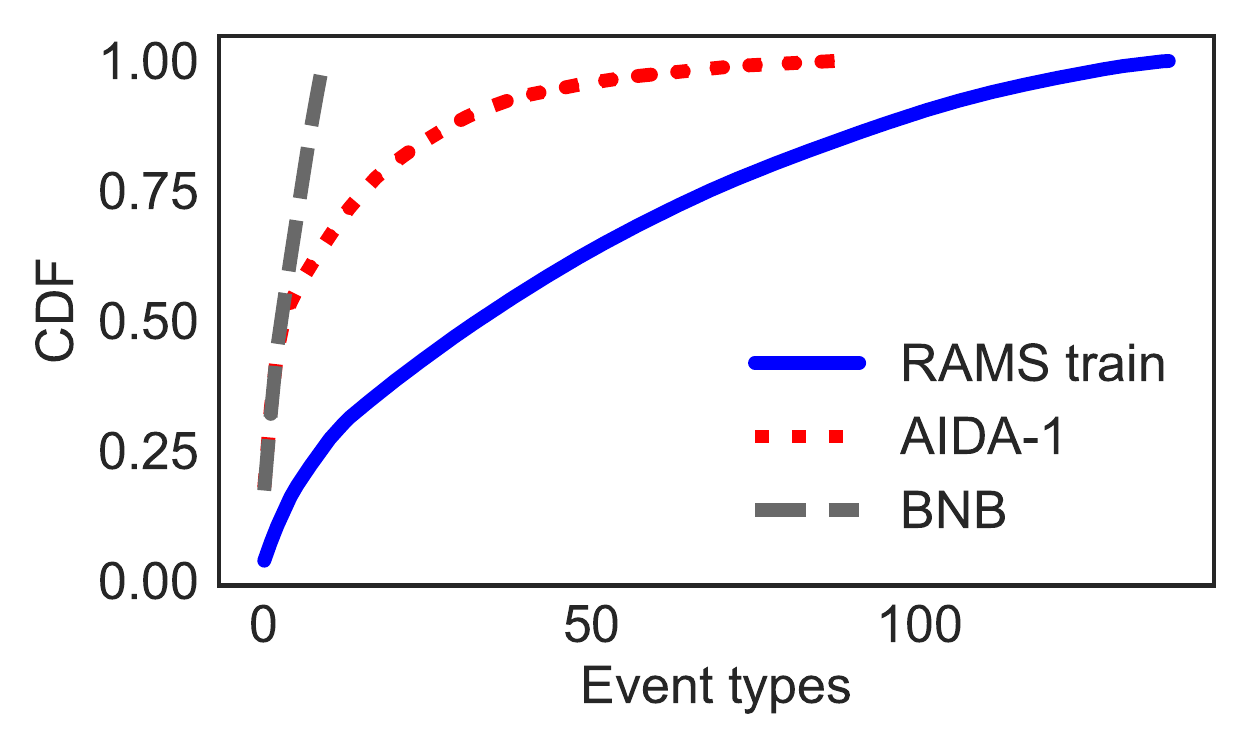}
\caption{Comparison of frequency (top) and amount of dataset covered (bottom) of event types sorted by decreasing frequency. \datasetname{} has more annotations for a more diverse set of event types than do AIDA Phase~1 and Beyond NomBank.}
\label{fig:type-coverage}
\end{figure}

\begin{figure}[t]
\includegraphics[width=0.9\linewidth]{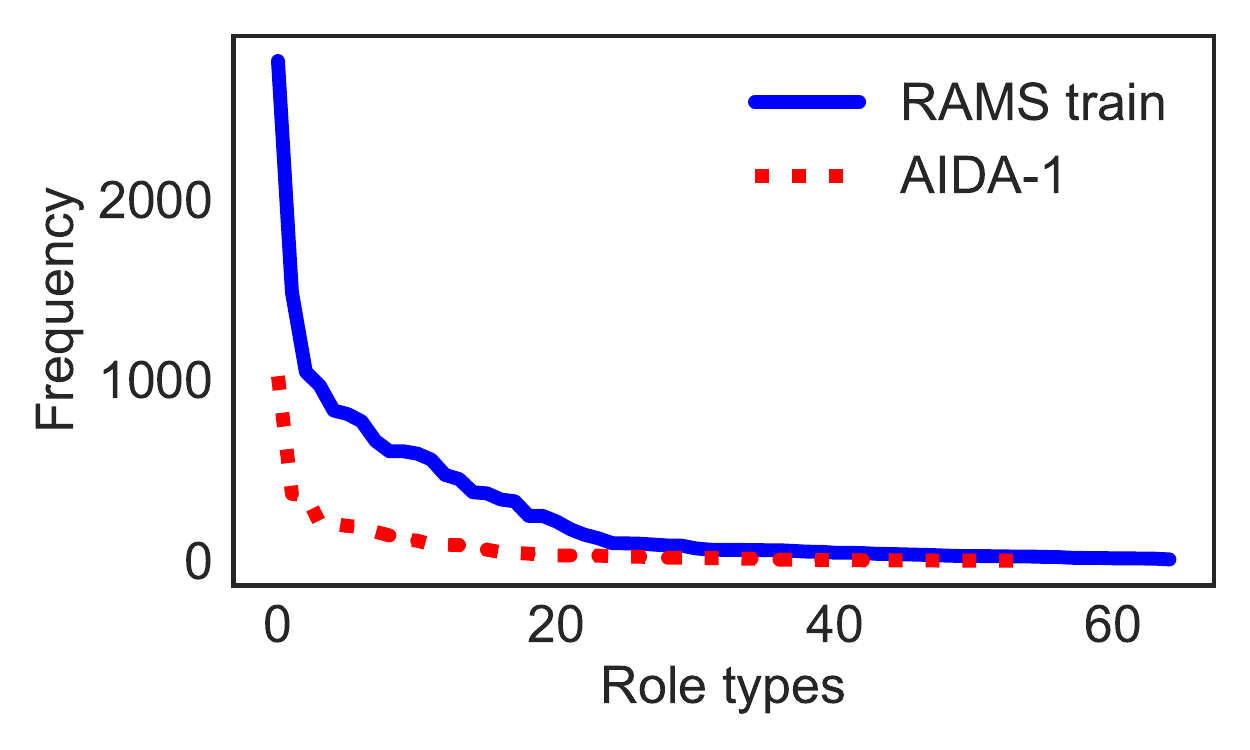}
\includegraphics[width=0.9\linewidth]{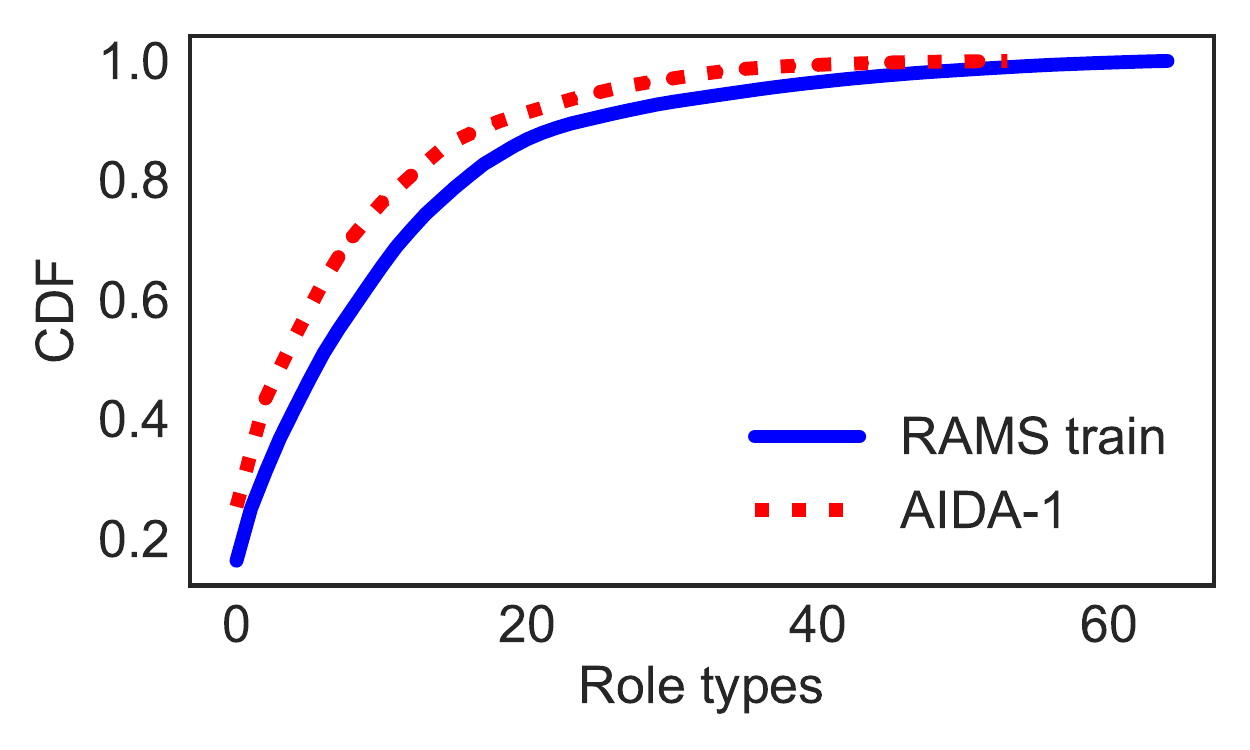}
\caption{Comparison of frequency (top) and amount of dataset covered (bottom) of roles sorted by decreasing frequency. \datasetname{} has more annotations for a more diverse set of role types than the AIDA Phase~1 data.}
\label{fig:role-coverage}
\end{figure}

Event type and role type coverage are shown in \autoref{fig:type-coverage} and \autoref{fig:role-coverage}. \autoref{fig:type-coverage} illustrates that \datasetname{} contains more annotations for a larger set of event types than does AIDA-1. In addition, the distribution of annotations in \datasetname{} is less skewed (more entropic) than in AIDA-1, in that in order to cover a given percentage of the dataset, more event types must be considered in \datasetname{} than in AIDA-1. \autoref{fig:role-coverage} shows a similar pattern for role type coverage.

\begin{figure}
\includegraphics[scale=0.45]{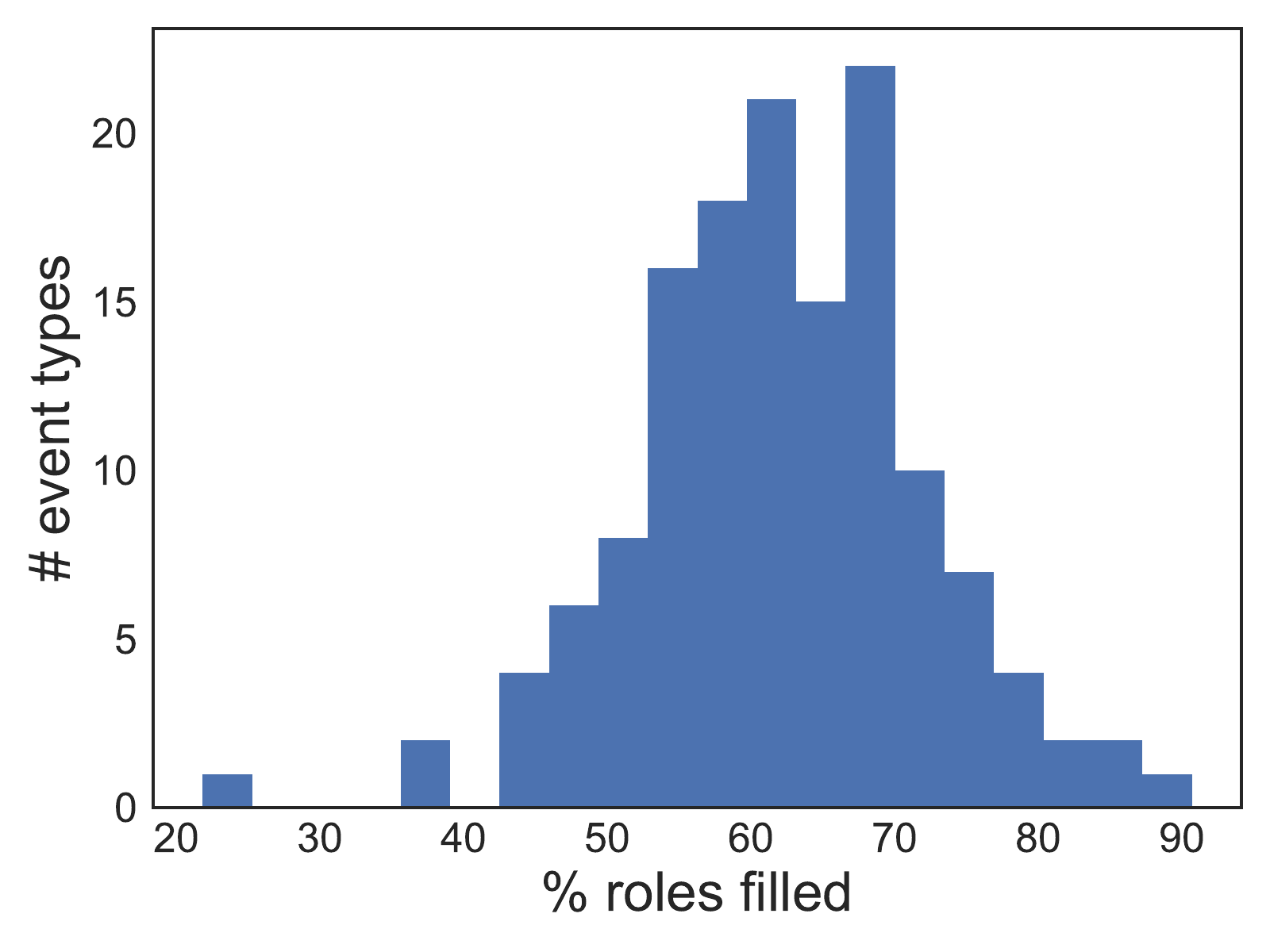}
\caption{Number of event types for which a given percentage of roles are filled in \datasetname{} train set.}
\label{fig:eventrole-coverage}
\end{figure}

\autoref{fig:eventrole-coverage} shows role coverage per event type, a measure of how much of each event type's role set is annotated on average. Role coverage per event type is calculated as the average number of filled roles per instance of the event type divided by the number of roles specified for that event type by the ontology. For the \datasetname{} training set, the 25\textsuperscript{th} percentile is 55.6\%, the 50\textsuperscript{th} percentile is 61.9\%, and the 75\textsuperscript{th} percentile is 68.6\% coverage.

\section{RAMS Hyperparameters}
\label{appendix:rams}

\begin{table}[t]
    \centering
    \small
    \begin{tabular}{cc|c}
    \toprule
    \multicolumn{2}{c}{Hyperparameter} &  Value \\
    \midrule
    \multirow{2}{*}{Embeddings} & role size & 50 \\
    & feature ($\phi_l$) size & 20 \\
    \multirow{3}{*}{LSTM} & size &  200 \\
    & layers & 3 \\
    & dropout & 0.4 \\
    \multirow{2}{*}{argument ($\text{F}_A$)} & size & 150 \\
    & layers & 2 \\
    \multirow{2}{*}{event-role ($\text{F}_{E,R}$)} & size & 150 \\
    & layers & 2 \\
    \multirow{1}{*}{$\text{F}_{\tilde{a}}$ (Eqn. \ref{eqn:trigrole})} & layers & 2 \\
    \multirow{2}{*}{arg-role ($\text{F}_{A,R}$)} & size & 150 \\
    & layers & 2 \\
    \multirow{2}{*}{$\text{F}_l$} & size & 150\\
    & layers & 2 \\
    \multirow{3}{*}{distance FFNN} & size & 150\\
    & layers & 2 \\
    & \# buckets & 10 \\
    \multirow{1}{*}{Pruning} & $k$ & 10\\
    \multirow{2}{*}{Memory Limits} & training doc size & 1000 \\
    & batch size & 1 \\
    \multirow{4}{*}{Training} & learning rate & 0.001 \\
    & decay & $\frac{0.999}{100 \text{~steps}}$ \\
    & patience & 10 \\
    \bottomrule
    \end{tabular}
    \caption{Hyperparameters of the model trained on RAMS. Sizes of learned weights that are omitted from the table can be determined from these hyperparameters. As the argument spans are given to the model in our experiments, we skip the first pass of pruning. We do not clip gradients.}
    \label{tab:rams_hparams}
\end{table}

\autoref{tab:rams_hparams} lists the numerical hyperparameters shared by all models discussed in this paper. Models may ignore some link score components if they were found to be unhelpful during our sweep of \autoref{eqn:link} and \autoref{eqn:big}. For our model, we learn a linear combination of the top layers (9, 10, 11, 12) of BERT-base cased, while we use the middle layers (6, 7, 8, 9) for the 6--9 ablation. For ELMo, we use all three layers and encode each sentence separately. We apply a lexical dropout of 0.5 to these embeddings.

In our best model, we use learned bucketed distance embeddings \cite{lee-etal-2017-end}. These embeddings are scored as part of $\phi_c$ in computing $s_c(e,a)$ in \autoref{eqn:link} and are also scored as a part of $\phi_\link$ in $s_\link$ (\autoref{eqn:big}). Since span boundaries are given in our primary experiments, we do not include a score $s_A$ or $s_E$ in $s_c$. Our best model uses both $s_{A, R}$ and $s_{l}(a, \tilde{a}_{e, r})$ in \autoref{eqn:link}. These features were chosen as the result of a sweep over possible features, with other ablations reported in \autoref{tab:ramsablations}.

We adopt the span embedding approach by \citet{lee-etal-2017-end}, which uses character convolutions (50 8-dimensional filters of sizes 3, 4, and 5) and 300-dimensional GloVe embeddings. The default dropout applied to all connections is 0.2. We optimize using Adam \cite{DBLP:journals/corr/KingmaB14} with patience-based early stopping, resulting in the best checkpoint after 19 epochs (9 hours on an NVIDIA 1080Ti), using \F{} as the evaluation metric.

Hyperparameters for the condition with distractor candidate arguments are the same as those in~\autoref{tab:rams_hparams}. For the condition with no given argument spans, we consider all intrasentential spans up to 5 tokens in length. We include the score of each candidate argument span when pruning to encourage the model to keep correct spans. We modify hyperparameters in \autoref{tab:rams_hparams} to prune less aggressively, setting $k=100$ and $\lambda_A=1.0$ (defined in \S\ref{sec:arch}).

\section{Full Role Confusion and Similarity Matrices}
\label{appendix:matrices}

\begin{figure*}[t]
\centering
\includegraphics[width=0.95\linewidth]{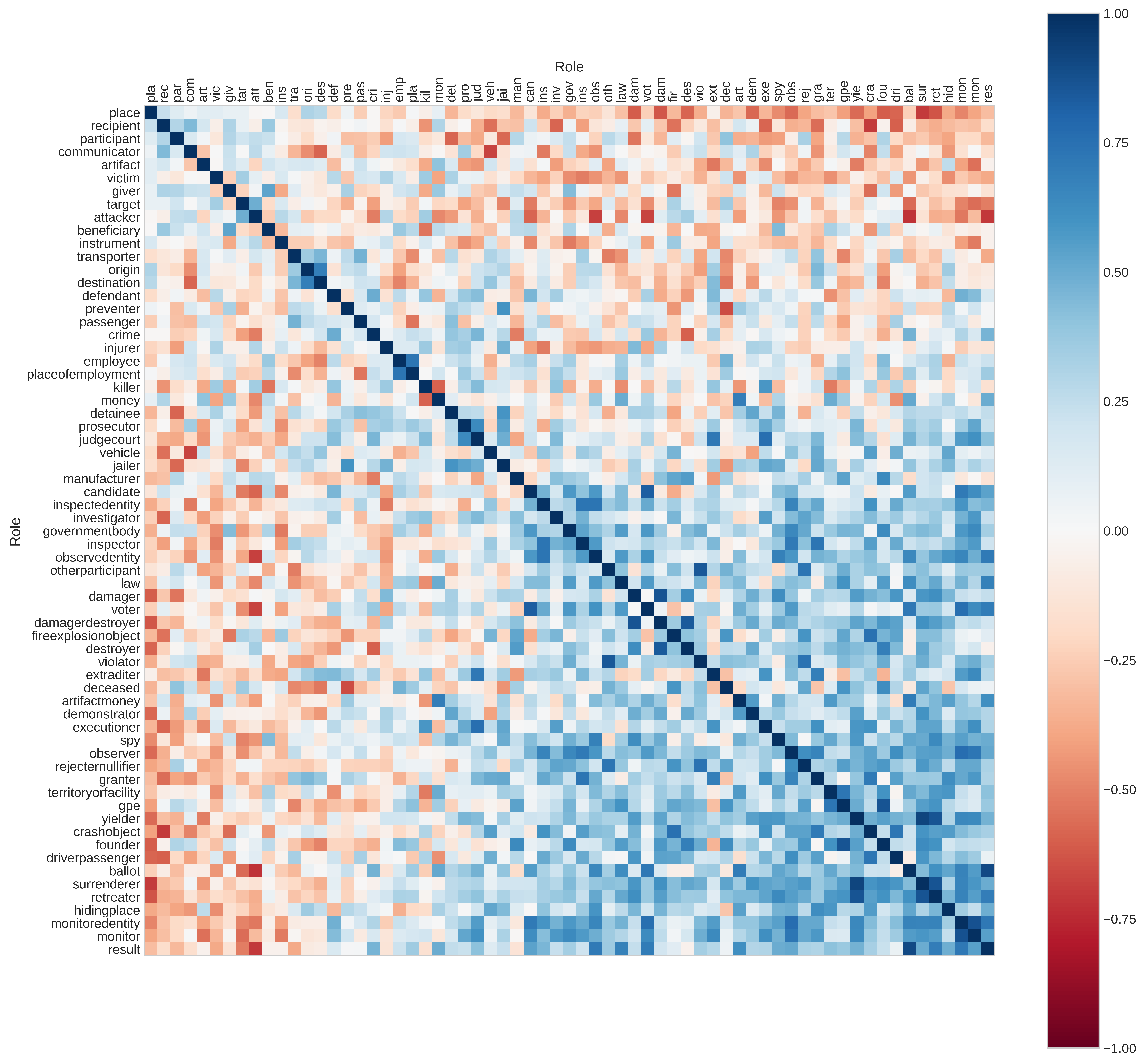}
\caption{Full version of \autoref{fig:embsim}, showing cosine similarity between role embeddings. Best viewed in color.}
\label{fig:embsim:full}
\end{figure*}

\begin{figure*}[t]
\centering
\includegraphics[width=0.95\linewidth]{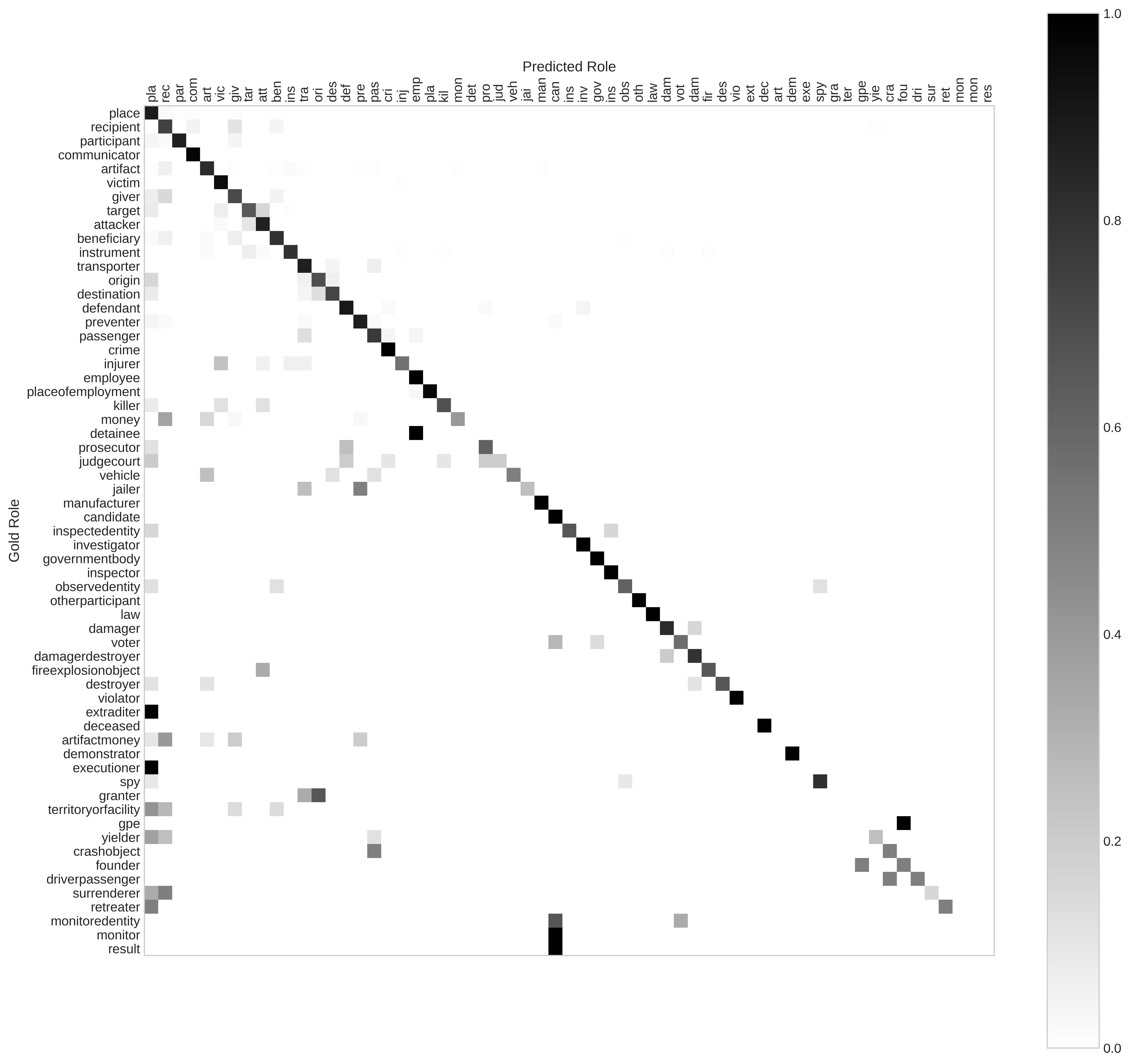}
\caption{Full version of \autoref{fig:embsim}, showing row-normalized confusion between roles. Note that roles not predicted at all would result in empty rows and so are omitted from the table.}
\label{fig:confusion:full}
\end{figure*}

\autoref{fig:embsim:full} shows the similarity between all 65 role embeddings, while \autoref{fig:confusion:full} visualizes all the errors made by the model on the development set. These are expansions of the per-role results from \S\ref{sec:rams:analysis}. 

Since argument linking is not a one-to-one labeling problem, we need to perform a modified procedure for visualizing a confusion matrix. For example, an argument span may take on multiple roles for the same event. To compute the errors, we first align the correct prediction(s) and subsequently compute the errors for the remaining gold and predicted label(s). For example, if the correct set of roles is \{\ont{destination, origin}\} and the model predicts \{\ont{origin, place}\}, then we only mark \ont{place} as an error for \ont{destination}.

\section{AIDA Phase~1}
\label{appendix:aida}

\subsection{Data Processing}
We filter and process the AIDA-1 Practice and Eval data in the following way. Because annotations are available for only a subset of the documents in AIDA-1, we consider only the documents that have textual event triggers. We then take from this set only the English documents, which, due to noisy language ID in the original annotations, were selected by manual inspection of the first 5 sentences of each document by one of the authors of this work.

In addition, the argument spans in each example are only those that participate in events. In other words, arguments of relations (that are not also arguments of events) are not included. Additionally, a document may contain multiple events, unlike in \datasetname{}.

The training and development set come from AIDA-1 Practice, and the test set comes from AIDA-1 Eval. As the AIDA-1 Eval documents are about different topics than the Practice documents are, we emulate the mismatch in topic distribution by using a development set that is about a different topic than the training set is. We use Practice topics \texttt{R103} and \texttt{R107} for training and \texttt{R105} for development because \texttt{R105} is the smallest of the three practice topics both by number of documents and by number of annotations. The test set consists of all 3 topics (\texttt{E101}, \texttt{E102}, \texttt{E103}) from the (unsequestered) Eval set. After the filtering process described above, we obtain a training set of 46 documents, a development set of 17 documents, and a test set of 69 documents. There are 389 events in the training set, and the training documents have an average length of 50 sentences.

\subsection{Hyperparameters}
We use the same hyperparameters as the best model for \datasetname{}, shown in \autoref{tab:rams_hparams}.

\subsection{Pre-training on \datasetname{}}
\begin{table}
\small
\centering
\begin{tabular}{lcccc}
\toprule
Strategy & Dev. \F{} & P & R & \F{} \\
\midrule
No pre-training & 25.0 & 36.6 & 12.9 & 19.1 \\ 
No pre-training\textsuperscript{TCD} & 27.1 & 53.5 & 11.0 & 18.2 \\
RAMS pre-training & 34.1 & 43.9 & \best{16.9} & 24.4 \\
RAMS pre-training\textsuperscript{TCD} & \best{34.2} & \best{62.5} & 15.4 & \best{24.8} \\
\bottomrule
\end{tabular}
\caption{P(recision), R(ecall), and \F{} on AIDA-1 English development and test data. TCD designates the use of ontology-aware type-constrained decoding.}
\label{tab:aidaresults}
\end{table} 

Both the models with and without pre-training on \datasetname{} were trained on AIDA-1 for 100 epochs with an early-stopping patience of 50 epochs using the same hyperparameters as the best \datasetname{} model. All parameters were updated during fine-tuning (none were frozen). The vocabulary of the pre-trained model was not expanded when trained on AIDA-1.

The models' lower performance on AIDA-1 than on \datasetname{} may be in part explained by the presence of distractors in AIDA-1. Moving from \datasetname{} (one trigger per example) to AIDA-1 (many triggers per example) introduces distractor ``negative'' links: an argument for one event might not participate in a different event in the same document. When given gold argument spans, a model learns from RAMS that every argument gets linked to the trigger, but there are many negative links in the AIDA-1 data, which the model must learn to not predict.

Full results are given in \autoref{tab:aidaresults}. Type-constrained decoding does not improve performance on AIDA-1 as much as it did in~\autoref{tab:ramsablations}, possibly because the AIDA-1 data often does not adhere to the multiplicity constraints of the ontology. For example, many \texttt{attack} events have more than one annotated \texttt{attacker} or \texttt{target}. Under TCD, correct predictions made in excess of what the ontology allows are deleted, hurting recall.

Interestingly, type-constrained decoding \emph{hurts} performance on AIDA-1 Eval when there is no pre-training. As discussed in \S\ref{sec:ramsexps}, type-constrained decoding tends to improve precision and lower recall. Despite the same behavior here, \F{} is nonetheless decreased.

We see similar behavior in this experiment to the \datasetname{} experiment involving distractor candidate arguments: low performance which is reduced further when using TCD.

\section{BNB Data Processing and Hyperparameters}
\label{appendix:bnb}

\subsection{Data Processing}

We use the data from \citet{gerber-chai-2012-semantic}.\footnote{ \url{http://lair.cse.msu.edu/projects/implicit_argument_annotations.zip}. Information about the data and its fields is available at \url{http://lair.cse.msu.edu/projects/implicit_annotations.html}.} We processed the data in the following way. The annotations were first aligned to text in the Penn Treebank. Because our model assumes that arguments are contiguous spans, we then manually merged all ``split'' arguments, which with one exception were already contiguous spans of text. For the one split argument that was not a contiguous span, we replaced it with its maximal span.\footnote{The instance is a quote broken by speaker attribution, where the split argument consists of the two halves of the quote. This example appears in our training set.} We then removed special parsing tokens such as ``trace'' terminals from the text and realigned the spans. While BNB gives full credit as long as one argument in each argument ``cluster'' is found, our training objective assumes one argument per role. We therefore automatically reduced each argument cluster to a singleton set containing the argument closest to the trigger. This reformulation of the problem limits our ability to compare to prior work.

Once all the data had been processed, we created training, development, and test splits. To avoid leaking information across splits, we bucketed examples by document and randomly assigned documents to the splits so that the splits contained instances in the proportions 80\% (train), 10\% (dev), and 10\% (test).

\subsection{Hyperparameters}
We use the same hyperparameters as the best model for \datasetname{}, shown in \autoref{tab:rams_hparams}.

\section{GVDB Hyperparameters and Additional Results}
\label{appendix:gvdb}

The entire GVDB corpus consists of 7,366 articles. We exclude articles that do not have a reliable publication date or lack annotated spans for the roles we are interested in. Additionally, a buffer of 100 articles spanning roughly one week between the dev and test set is discarded, limiting the possibility of events occurring in both the development and test sets. We also filter out spans whose start and end boundaries are in different sentences, as these are unlikely to be well-formed argument spans. For evaluation, a slot's value is marked as correct under the \textit{strict} setting if any of the predictions for that slot match the string of the correct answer exactly, while an \textit{approximate} match is awarded if either a prediction contains the correct answer or if the correct answer contains the predicted string. The approximate setting is necessary due to inconsistent annotations (e.g., omitting first or last names). 

\begin{table}[t]
    \centering
    \small
    \begin{tabular}{cc|c}
    \toprule
    \multicolumn{2}{c}{Hyperparameter} &  Value \\
    \midrule
    \multirow{2}{*}{Embeddings} & role size & 50 \\
    & feature ($\phi_l$) size & 20 \\
    \multirow{3}{*}{LSTM} & size &  200 \\
    & layers & 3 \\
    & dropout & 0.4 \\
    \multirow{2}{*}{argument ($\text{F}_A$)} & size & 150 \\
    & layers & 2 \\
    \multirow{2}{*}{event-role ($\text{F}_{E,R}$)} & size & 150 \\
    & layers & 2 \\
    \multirow{1}{*}{$\text{F}_{\tilde{a}}$ (Eqn. \ref{eqn:trigrole})} & layers & 2 \\
    \multirow{2}{*}{$\text{F}_l$} & size & 150\\
    & layers & 2 \\
    \multirow{2}{*}{positional FFNN} & size & 150\\
    & layers & 2 \\
    & \# buckets & 10 \\
    \multirow{2}{*}{Pruning} & $\lambda_A$ & 0.8 \\
    & $k$ & 45\\
    \multirow{3}{*}{Memory Limits} & training doc size & 600 \\
    & span width & 15 \\
    & batch size & 1 \\
    \multirow{4}{*}{Training} & learning rate & 0.0005 \\
    & decay & $\frac{0.999}{200 \text{~steps}}$ \\
    & patience & 20 \\
    & gradient clipping & 10.0 \\
    \bottomrule
    \end{tabular}
    \caption{Hyperparameters of the model trained on GVDB.}
    \label{tab:gvdb_hparams}
\end{table}

We experiment with the feature-based version of BERT-base and with ELMo as our contextualized encoder. \autoref{tab:gvdb_hparams} lists the numerical hyperparameters for this model. Since there is only one event per document and no explicit trigger, $e$ is represented by a span embedding of the full document. We use the top four layers (9--12) of BERT-base cased (all three layers for ELMo) with a lexical dropout of 0.5. Everywhere else, we apply a dropout of 0.4. We train with the Adam optimizer \cite{DBLP:journals/corr/KingmaB14} and use patience-based early stopping. Our best checkpoint was after 8 epochs (roughly 9 hours on a single NVIDIA 1080Ti). Even though the official evaluation is string based, we used a span-based micro \F{} metric for early stopping.

For this model, $\phi_l$ corresponds to a learned (bucketed) positional embedding of the argument span (i.e., distance from the start of the document). In computing the coarse score, we omit $\phi_c$. When computing \autoref{eqn:link}, we omit $s_{A, R}$ but keep all other terms in \autoref{eqn:link}. We adopt the character convolution of 50 8-dimensional filters of window sizes 3, 4, and 5 \cite{lee-etal-2017-end}.

With the same hyperparameters and feature choices, we perform an identical evaluation using ELMo instead of BERT. As the original documents are not tokenized, we use \texttt{SpaCy 2.1.4} for finding sentence boundaries and tokenization. The complete list of annotated fields are \textsc{victim} (name, age, race), \textsc{shooter} (name, age, race), \textsc{location} (specific location\footnote{For example, a park or a laundromat.} or city), \textsc{time} (time of day or clock time) and \textsc{weapon} (weapon type, number of shots fired). While \citet{pavlick-etal-2016-gun} only make predictions for {\sc Victim.Name}, {\sc Shooter.Name}, {\sc Location.(City$|$Location)}, {\sc Time.(Time$|$Clock)}, and {\sc Weapon.Weapon}, we perform predictions over all annotated span-based fields. The full results for both BERT and ELMo are reported in \autoref{tab:gvdb_results} and \autoref{tab:gvdb_elmo_results}, respectively. BERT generally improves over ELMo across the board, but not by a sizeable margin. Despite the inability to directly compare, we nonetheless present a stronger and more comprehensive baseline for future work with GVDB.



\begin{table*}[t]
    \centering
    \small
    
    \begin{tabular}{llcccccclcccccc}
    \toprule
    \multicolumn{2}{c}{Field} & \multicolumn{6}{c}{Strict} & & \multicolumn{6}{c}{Partial} \\
    & & \multicolumn{3}{c}{Baseline} & \multicolumn{3}{c}{Us} & & \multicolumn{3}{c}{Baseline} & \multicolumn{3}{c}{Us}\\
    & & P & R & F1 & P & R & F1 & & P & R & F1 & P & R & F1 \\
    \midrule
    \multirow{3}{*}{\textsc{victim}} & Name & \textit{10.2} & \textit{8.5} & \textit{9.3} & 61.2 & 63.3 & 62.2 & & \textit{59.5} & \textit{49.6} & \textit{54.1} & 68.4 & 70.9 & 69.6 \\
    & Age & -- & -- & -- & 19.4 & 24.2 & 21.5 & & -- & -- & -- & 67.3 & 84.1 & 74.8 \\
    & Race & -- & -- & -- & 75.5 & 74.1 & 74.8 & & -- & -- & -- & 75.5 & 74.1 & 74.8\\
    \noalign{\vskip 2mm}
    \multirow{3}{*}{\textsc{shooter}} & Name & \textit{5.8} & \textit{3.9} & \textit{4.7} & 55.3 & 51.1 & 53.1 & & \textit{30.2} & \textit{20.1} & \textit{24.1} & 60.2 & 55.6 & 57.8\\
    & Age & -- & -- & -- & 34.1 & 32.6 & 33.3 & & -- & -- & -- & 69.0 & 65.9 & 67.4 \\
    & Race & -- & -- & -- & 72.7 & 55.2 & 62.7 & & -- & -- & -- & 81.8 & 62.1 & 70.6\\
    \noalign{\vskip 2mm}
    \multirow{2}{*}{\textsc{location}} & City & \multirow{2}{*}{\textit{19.9}} &\multirow{2}{*}{\textit{8.8}} & \multirow{2}{*}{\textit{12.2}} & 67.4 & 66.2 & 66.8 & & \multirow{2}{*}{\textit{30.8}}  & \multirow{2}{*}{\textit{13.6}}  & \multirow{2}{*}{\textit{18.9}}  & 72.2 & 70.9 & 71.5 \\
    & Location &  & &  & 36.1 & 33.8 & 34.9 & & & & & 65.4 & 61.2 & 63.3 \\
    \noalign{\vskip 2mm}
    \multirow{2}{*}{\textsc{time}} & Time & \multirow{2}{*}{\textit{69.3}} & \multirow{2}{*}{\textit{66.9}} & \multirow{2}{*}{\textit{68.1}} & 57.2 & 69.7 & 62.9 & & \multirow{2}{*}{\textit{70.5}} & \multirow{2}{*}{\textit{68.1}} & \multirow{2}{*}{\textit{69.3}} & 63.2 & 76.9 & 69.4 \\
    & Clock &  &  & & 44.0 & 47.6 & 45.7 & &  &  & & 84.0 & 90.8 & 87.2  \\\noalign{\vskip 2mm}
    \multirow{2}{*}{\textsc{weapon}} & Weapon & \textit{2.1} & \textit{0.7} & \textit{1.1} & 33.3 & 31.7 & 32.5 & & \textit{36.8} & \textit{11.8} & \textit{17.9} & 50.9 & 48.3 & 49.6 \\
    & Num Shots & -- & -- & -- & 40.6 & 11.2 & 17.6 & & -- & -- & -- & 62.5 & 17.2 & 27.0 \\
    \bottomrule
    \end{tabular}
    \caption{P(recision), R(ecall), and \F{} on event-based slot filling (GVDB) using BERT as the document encoder. Due to the different data splits and evaluation conditions, the results are not directly comparable to the baseline~\cite{pavlick-etal-2016-gun}, which is provided only for reference. Fields that were aggregated in the baseline are predicted separately in our model. `--' indicates result is not reported in the baseline.}
    \label{tab:gvdb_results}
\end{table*}
\begin{table*}[t]
    \centering
    \small
    
    \begin{tabular}{llcccccclcccccc}
    \toprule
    \multicolumn{2}{c}{Field} & \multicolumn{6}{c}{Strict} & & \multicolumn{6}{c}{Partial} \\
    & & \multicolumn{3}{c}{Baseline} & \multicolumn{3}{c}{Us} & & \multicolumn{3}{c}{Baseline} & \multicolumn{3}{c}{Us}\\
    & & P & R & F1 & P & R & F1 & & P & R & F1 & P & R & F1 \\
    \midrule
    \multirow{3}{*}{\textsc{victim}} & Name & \textit{10.2} & \textit{8.5} & \textit{9.3} & 56.2 & 56.8 & 56.5 & & \textit{59.5} & \textit{49.6} & \textit{54.1} & 62.7 & 63.3 & 63.0 \\
    & Age & -- & -- & -- & 29.5 & 33.9 & 31.6 & & -- & -- & -- & 64.4 & 74.0 & 68.9 \\
    & Race & -- & -- & -- & 73.2 & 75.9 & 74.5 & & -- & -- & -- & 75.0 & 77.8 & 76.4\\
    \noalign{\vskip 2mm}
    \multirow{3}{*}{\textsc{shooter}} & Name & \textit{5.8} & \textit{3.9} & \textit{4.7} & 53.7 & 60.2 & 56.7 & & \textit{30.2} & \textit{20.1} & \textit{24.1} & 56.4 & 63.2 & 59.6\\
    & Age & -- & -- & -- & 27.3 & 31.8 & 29.4 & & -- & -- & -- & 53.2 & 62.1 & 57.3 \\
    & Race & -- & -- & -- & 55.9 & 65.5 & 60.3 & & -- & -- & -- & 58.8 & 69.0 & 63.5\\
    \noalign{\vskip 2mm}
    \multirow{2}{*}{\textsc{location}} & City & \multirow{2}{*}{\textit{19.9}} &\multirow{2}{*}{\textit{8.8}} & \multirow{2}{*}{\textit{12.2}} & 59.1 & 61.1 & 60.1 & & \multirow{2}{*}{\textit{30.8}}  & \multirow{2}{*}{\textit{13.6}}  & \multirow{2}{*}{\textit{18.9}}  & 64.1 & 66.2 & 65.1 \\
    & Location &  & &  & 36.6 & 34.7 & 35.6 & & & & & 59.1 & 56.0 & 57.5 \\
    \noalign{\vskip 2mm}
    \multirow{2}{*}{\textsc{time}} & Time & \multirow{2}{*}{\textit{69.3}} & \multirow{2}{*}{\textit{66.9}} & \multirow{2}{*}{\textit{68.1}} & 57.7 & 64.7 & 61.0 & & \multirow{2}{*}{\textit{70.5}} & \multirow{2}{*}{\textit{68.1}} & \multirow{2}{*}{\textit{69.3}} & 64.5 & 72.4 & 68.2 \\
    & Clock &  &  & & 44.6 & 45.8 & 45.2 & &  &  & & 83.5 & 85.6 & 84.5  \\\noalign{\vskip 2mm}
    \multirow{2}{*}{\textsc{weapon}} & Weapon & \textit{2.1} & \textit{0.7} & \textit{1.1} & 32.7 & 26.7 & 29.4 & & \textit{36.8} & \textit{11.8} & \textit{17.9} & 44.9 & 36.7 & 40.4 \\
    & Num Shots & -- & -- & -- & 23.3 & 18.1 & 20.4 & & -- & -- & -- & 42.2 & 32.8 & 36.9 \\
    \bottomrule
    \end{tabular}
    \caption{P(recision), R(ecall), and \F{} on event-based slot filling (GVDB) using ELMo at the sentence level. On average, the performance is outperformed by BERT.}
    \label{tab:gvdb_elmo_results}
    
\end{table*}

\end{document}